\documentclass[10pt,twocolumn,letterpaper]{article}
\PassOptionsToPackage{numbers,sort&compress}{natbib}

\usepackage[pagenumbers]{cvpr} 

\usepackage[title]{appendix}
\usepackage{multirow}
\usepackage{amsmath}

\usepackage[dvipsnames]{xcolor}
\usepackage{amsmath}
\usepackage{amssymb}
\usepackage{amsthm}
\usepackage{soul}
\usepackage{bbm}
\usepackage[most]{tcolorbox}

\definecolor{dark_green}{rgb}{0, 0.5, 0}
\definecolor{dark_blue}{rgb}{0, 0, 0.5}

\definecolor{cvprblue}{rgb}{0.21,0.49,0.74}
\usepackage[pagebackref,breaklinks,colorlinks,citecolor=cvprblue]{hyperref}
\usepackage{nicefrac}

\def\confName{CVPR}
\def\confYear{2024}

\newcommand{\parnobf}[1]{\vspace{0.5mm} \par \noindent {\bf {#1}.}}
\newcommand{\parnoit}[1]{\vspace{0.5mm} \par \noindent {\it {#1}.}}

\definecolor{ibm1}{HTML}{648FFF}
\definecolor{ibm2}{HTML}{DC267F}
\definecolor{ibm3}{HTML}{FE6100}
\definecolor{ibm4}{HTML}{FFB000}
\definecolor{ibm5}{HTML}{785EF0}
\definecolor{ibm6}{HTML}{88CCEE}

\newcommand{\SOT}{{\mathrm{SO}(3)}}
\newcommand{\SET}{{\mathrm{SE}(3)}}

\renewcommand{\paragraph}[1]{\vspace{2pt} \noindent \textbf{#1}}

\newcommand{\gB}{\mathbf{g}}

\newcommand{\pB}{\mathbf{p}}
\newcommand{\qB}{\mathbf{q}}

\newcommand{\tB}{\mathbf{t}}

\newcommand{\wB}{\mathbf{w}}

\newcommand{\EB}{\mathbf{E}}

\newcommand{\GB}{\mathbf{G}}
\newcommand{\HB}{\mathbf{H}}

\newcommand{\MB}{\mathbf{M}}

\newcommand{\RB}{\mathbf{R}}

\newcommand{\TB}{\mathbf{T}}

\title{FAR: Flexible, Accurate and Robust 6DoF Relative Camera Pose Estimation}

\author{Chris Rockwell$^1$ \qquad Nilesh Kulkarni$^1$ \qquad Linyi Jin$^1$ \\ 
Jeong Joon Park$^1$ \qquad Justin Johnson$^1$ \qquad David F. Fouhey$^2$ \\
University of Michigan$^1$ \qquad New York University$^2$ 
}

\begin{document}
\maketitle
\begin{abstract}
\vspace{-0.3em}
Estimating relative camera poses between images has been a central problem in computer vision. Methods that find correspondences and solve for the fundamental matrix offer high precision in most cases. Conversely, methods predicting pose directly using neural networks are more robust to limited overlap and can infer absolute translation scale, but at the expense of reduced precision. We show how to combine the best of both methods; our approach yields results that are both precise and robust, while also accurately inferring translation scales. At the heart of our model lies a Transformer that (1) learns to balance between solved and learned pose estimations, and (2) provides a prior to guide a solver. A comprehensive analysis supports our design choices and demonstrates that our method adapts flexibly to various feature extractors and correspondence estimators, showing state-of-the-art performance in 6DoF pose estimation on Matterport3D, InteriorNet, StreetLearn, and Map-free Relocalization. Project page: \href{https://crockwell.github.io/far/}{https://crockwell.github.io/far/}
\end{abstract}
\vspace{-1.2em}

\section{Introduction}
\label{sec:intro}

\begin{figure}
\begin{center}
\includegraphics[width=\columnwidth]{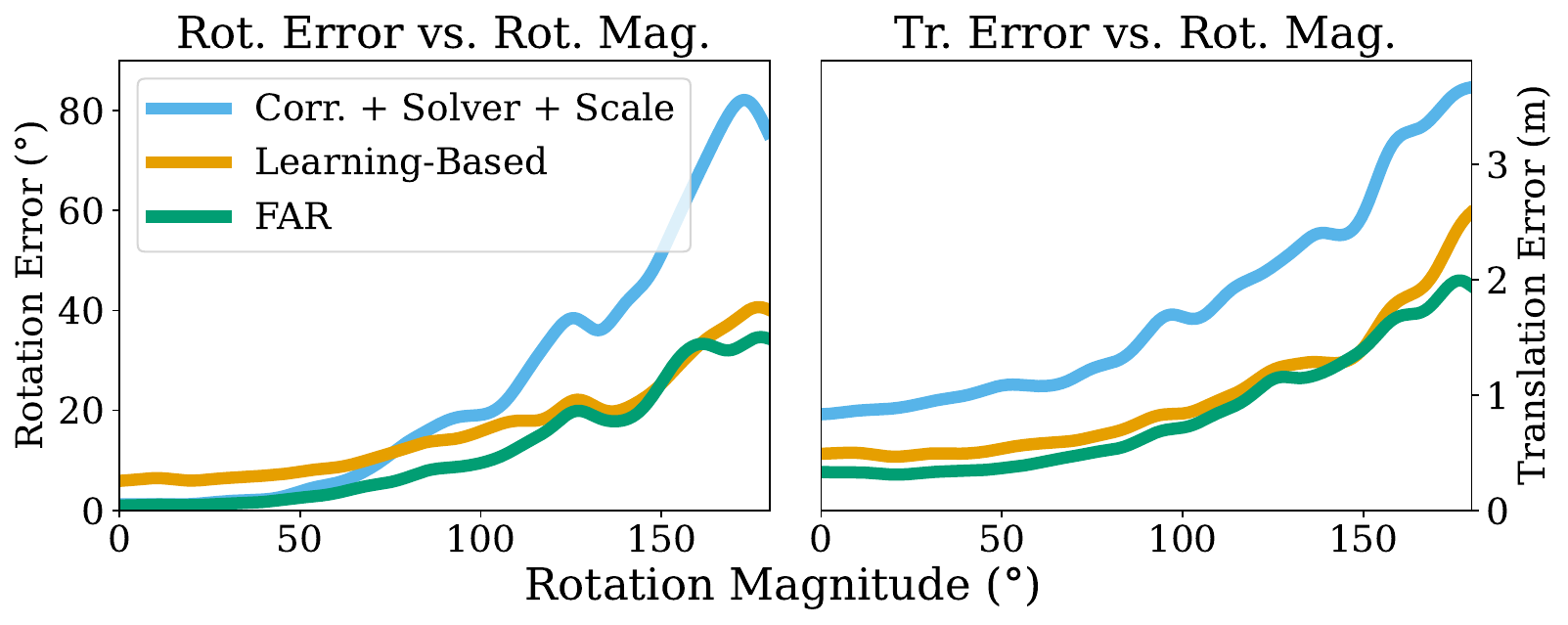}
\vspace{-1em}
\caption{\textbf{Precise and Robust 6DoF Pose Estimation}. Correspondence Estimation + Solver methods (here LoFTR~\cite{sun2021loftr}, RANSAC~\cite{Fischler81}) produce precise outputs for moderate rotations, but are not robust to large rotations (left), and cannot produce translation scale.
Learning-based methods (here LoFTR with 8-Point ViT~\cite{rockwell2022} head) produce scale (right) and are more robust, but lack precision (left).
FAR leverages both for precise and robust prediction, including scale.
}
\label{fig:teaser}
\end{center}
\vspace{-2em}
\end{figure}

Relative camera pose estimation is a fundamental problem in computer vision~\cite{hartley2003multiple}, with applications in augmented reality~\cite{lin2021barf,lai2021video,jiang2022few}, robotics~\cite{mur2015orb,schonberger2016structure,yu2018ds}, and autonomous driving~\cite{geiger2012we,bresson2017simultaneous}. 
One recent line of work learns to estimate correspondences then solve for pose~\cite{sun2021loftr,sarlin2020superglue,lindenberger2023lightglue,edstedt2023dkm,ni2023pats}, often offering sub-degree errors.
Unfortunately, this framework tends to struggle when faced with large view change (Figure~\ref{fig:teaser}, left), and additionally cannot recover scale because it produces the Fundamental or Essential matrix. 
Another line of work learns to estimate pose directly~\cite{rockwell2022,cai2021extreme,wang2023posediffusion,lin2023relpose++,barroso2023two}, which is not as precise, but can be more robust and produces translation scale (Figure~\ref{fig:teaser}, left and right).

The proposed method builds upon both communities to produce a general method that is no worse than either of the options and often better than both.
Critically, it leverages learned correspondence predictions as input, and combines learned pose estimation with a solver to estimate 6DoF pose.
For this task, we purposefully select the Transformer, which can handle dense features or correspondences as input.
Put succinctly, the method is \textbf{F}lexible: agnostic to correspondence and feature backbone; \textbf{A}ccurate: matches the precision of correspondence-based methods; and \textbf{R}obust: builds upon the resilience of learned pose methods.

FAR enables learning-based and solver-based methods to improve each other.
Learned predictions are more robust than solver output, and are therefore used as a prior to bias the solver.
Improved solver output, which tends to be more precise than learned output when it succeeds, is then combined with Transformer predictions to form final output.
Predictions are combined via a weighting predicted by the Transformer,
 meaning the Transformer can learn to rely more upon either method depending on their effectiveness.

Figure~\ref{fig:analysis} analyzes FAR in practice, measuring error as a function of the number of good input correspondences.
With many correspondences, the solver is highly accurate, leaving little room for improvement from the prior.
As the number of correspondences drop, solver performance  degrades, but this can be alleviated meaningfully using the learned prior.
The learned weighting also contributes to robustness, and is plotted on the right: the Transformer primarily uses solver output if there are many correspondences, and more heavily uses the regressor when there are few correspondences (Figure~\ref{fig:analysis}, right).
The result is a method that does not sacrifice in the case of many correspondences, but has a large gain given few correspondences.

Experiments analyze FAR in detail across a number of scenarios and datasets.
First, we analyze theoretical robustness, beginning from ground truth correspondences and procedurally adding (1) noise and (2) outliers. 
We next evaluate the proposed method on four challenging datasets, spanning both indoors: Matterport3D and InteriorNet, and outdoors: StreetLearn and Map-free Relocalization.
Across settings, the proposed method typically outperforms, or occasionally matches, the state of the art.
We additionally analyze the components of FAR in ablations, and apply it to a variety of permutations of correspondence and feature estimation backbones.
We also study the impact of dataset size upon model behavior.

\section{Related Work}
\label{sec:related}

\parnobf{Learned Camera Pose Estimation}
Learned Camera Pose Estimation has recently made impressive progress. 
If many views are available, camera pose can be precisely refined during SLAM~\cite{teed2021droid,czarnowski2020deepfactors} or Visual Odometry~\cite{lai2023xvo,wang2021tartanvo,teed2022deep}. 
If fewer views of a scene are available, methods have become increasingly robust to e.g. large rotation~\cite{zhang2022relpose,sinha2023sparsepose,jiang2022few}. 

This paper focuses on the wide-baseline two-view setting, which has also strongly progresses~\cite{chen2021wide,cai2021extreme,en2018rpnet,yang2020extreme,yang2019extreme}.
Some of these methods also perform 3D reconstruction~\cite{agarwala2022planes,jin2021planar, qian2020associative3d,tan2023nope}.
We build off the 8-Point ViT \cite{rockwell2022}, a SOTA method for two-image 6DoF camera pose estimation. 

\parnobf{Correspondence Estimation}
Correspondence can be learned~\cite{huang2023adaptive,ni2023pats,chen2022aspanformer,Dusmanu2019CVPR,edstedt2023dkm,kloepferscenes} or attained using classical methods~\cite{lowe2004distinctive,rublee2011orb,bay2006surf}, including specialized for wide-baseline stereo~\cite{matas2004robust,mishkin2015wxbs,Pritchett98a}.
We use recent SOTA methods LoFTR~\cite{sun2021loftr} and SuperPoint+SuperGlue~\cite{detone2018superpoint,sarlin2020superglue}, but note FAR can readily adapt to alternative estimators.

\parnobf{Camera Pose Estimation from Correspondences}
Camera pose estimation from correspondences is a long standing~\cite{Fischler81} and still active problem~\cite{barath2019magsac}. Typically, algorithms use a robust estimator~\cite{barath2020magsac++,barath2019magsac,Fischler81} along with the 7-Point~\cite{larsson2018beyond} or 8-Point~\cite{hartley1997defense} algorithm to find the fundamental matrix, if intrinsics are unknown, or 5-Point algorithm~\cite{nister2004efficient} to find the essential matrix, if intrinsics are known. F or E can than be decomposed into RT (without translation scale) and estimating direction via triangulation and the chirality check. We assume known intrinsics and use RANSAC with the 5-Point algorithm, but our contributions are orthogonal to estimator.

\begin{figure}[t]
    \centering
    \includegraphics[width=\linewidth]{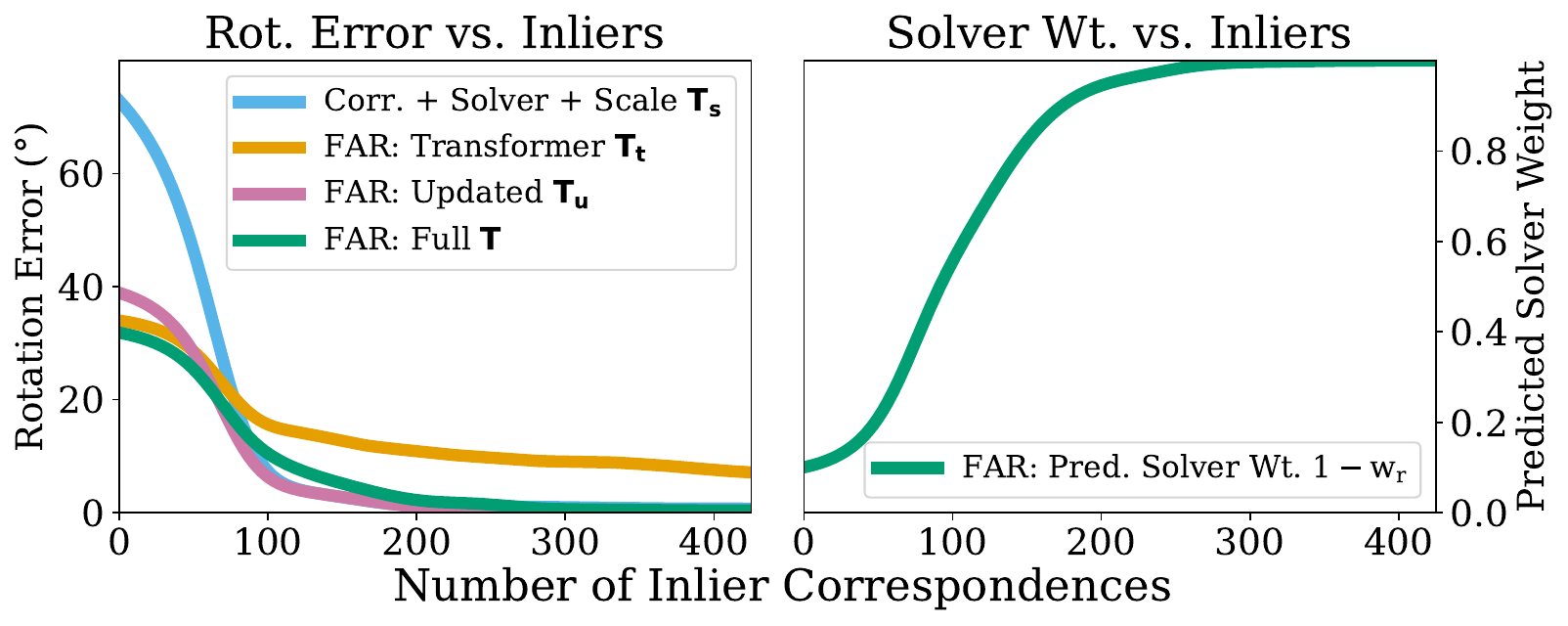}
    \vspace{-1.7em}
    \caption{\textbf{Combining Classical and Learned}.
    Left: Solver output is precise given many inliers, but is poor when few are available; Updated solver output via FAR's prior improves robustness significantly.
    FAR's Transformer is less precise but more robust.
    The full model fuses prior-guided Solver output and Transformer output for the best of both, giving more weight to the solver when many inliers are available (right).
    }
    \label{fig:analysis}
    \vspace{-1.3em}
\end{figure}

Recent work incorporates learnable elements into pose estimation from correspondence~\cite{ranftl2018deep}. 
Barroso \textit{et al.}~\cite{barroso2023two} learn to select from candidate essential matrices. 
Roessle and Nießner \cite{roessle2023end2end} use a differentiable 8-Point algorithm, while Wei \textit{et al.}~\cite{wei2023generalized,zhao2021progressive} use a differentiable robust estimator, to improve F via end-to-end training. 
DSAC~\cite{brachmann2017dsac} supervises a Scoring CNN to predict consensus and guide RANSAC. 
FAR is distinct from these works as its final output is a weighted combination of Solver Pose and Learned Pose.
This important difference allows FAR to predict scale and improve robustness to poor or limited correspondences (Figure~\ref{fig:analysis}).
GRelPose~\cite{khatib2022grelpose} predicts pose from correspondences, but does not use a solver, limiting precision. 

\section{Approach}
\label{sec:approach}

Our goal is to predict relative camera pose, including translation scale, from two overlapping images. 
This 6DoF pose can be parameterized as $\TB \in \SET$, consisting of $\RB \in \SOT$ and $\tB \in \mathbb{R}^3$.
We specifically focus on predicting translation scale, which cannot be solved for from correspondences alone, in order to enable real world applications e.g. 3D reconstruction and neural rendering.
The two-view case facilitates these applications on e.g. image collections.
We assume known camera intrinsics as they are generally available from modern devices~\cite{arnold2022map}.

\newcommand{\ig}[1]{\includegraphics[height=0.8em,width=0.8em]{figures/glyphs/#1.pdf}}

\definecolor{colortt}{HTML}{E69F00}
\definecolor{colorts}{HTML}{57B4E9}
\definecolor{colort}{HTML}{019E73}
\definecolor{colort1}{HTML}{F0E442}
\definecolor{colortp}{HTML}{BB4D89}
\definecolor{colorw}{HTML}{019E73}

\DeclareRobustCommand{\hltt}[1]{{\sethlcolor{colortt}\hl{#1}}}
\DeclareRobustCommand{\hlts}[1]{{\sethlcolor{colorts}\hl{#1}}}
\DeclareRobustCommand{\hlto}[1]{{\sethlcolor{colort1}\hl{#1}}}
\DeclareRobustCommand{\hltp}[1]{{\sethlcolor{colortp}\hl{#1}}}
\DeclareRobustCommand{\hlw}[1]{{\sethlcolor{colorw}\hl{#1}}}
\DeclareRobustCommand{\hlt}[1]{{\sethlcolor{colort}\hl{#1}}}

\begin{figure*}[t]
    \centering
    \includegraphics[width=\linewidth]{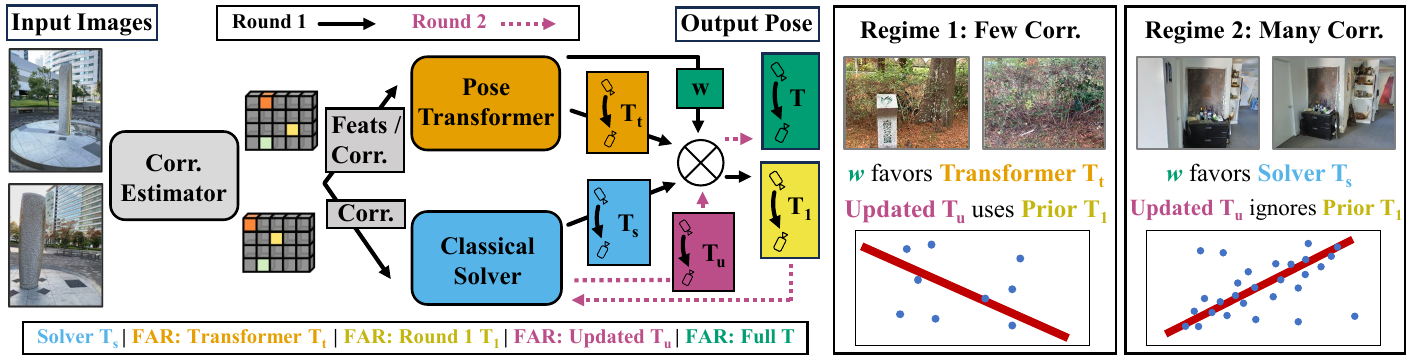}
    \vspace{-1.3em}
    \caption{\textbf{Overview}. 
    Given dense features and correspondences, FAR's Transformer produces camera poses (in square boxes \ig{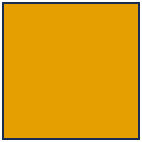}) through a transformer (round box \ig{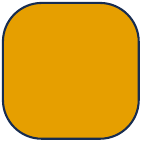}) and classical solver (round box \ig{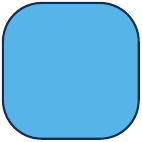}). In the first round, the solver produces a pose \hlts{$\TB_s$}. FAR's pose transformer averages this with its own prediction \hltt{$\TB_t$} via weight \hlw{$\wB$}, to yield the round 1 pose \hlto{$\TB_1$}.  \hlto{$\TB_1$} pose serves as a prior for the classic solver, which produces an updated pose \hltp{$\TB_u$}. This is combined with an additional estimate of \hltt{$\TB_t$} and weight \hlw{$\wB$} to produce the final result \hlt{$\TB$}.
    With few correspondences, $\TB_1$ helps solver output, while the network learns to weigh Transformer predictions more heavily; 
    with many correspondences, solver output is often good, so the network relies mostly on solver output.}
    \label{fig:method_overview}
    \vspace{-1.5em}
\end{figure*}

FAR fuses complimentary strengths from the two lines of pose estimation work: learned correspondence estimation followed by a robust solver, and end-to-end pose estimation.
Critically, it produces results that are no worse than either and often better than both. We design the method as flexible to be plugged-in to existing methods with minimal change, showing improved results in a variety of settings and datasets.
We outline FAR in Section~\ref{sec:approach_outline}, detail the learned network in Section~\ref{sec:approach_learned} and how we apply a prior to the solver in Section~\ref{sec:approach_solver}.

\subsection{Approach Outline}
\label{sec:approach_outline} 

Figure~\ref{fig:method_overview} shows an overview of the proposed approach. 
At the heart of the approach is the Pose Transformer (Sec~\ref{sec:approach_learned}) which takes in dense features and outputs (1) an estimate of 6DoF pose $\TB_t$ and (2) a relative weight $\wB$ of this prediction. 
$\TB_t$ is then combined with solver-estimated pose $\TB_s$ using weight $\wB$ to obtain pose estimate $\TB_1$.
$\TB_1$ is used as a prior for the solver, resulting in updated solver output $\TB_u$, which is combined with $\TB_t$ via $\wB$ to get the final output $\TB$.

This architecture enables the network to learn to behave differently depending on the data regime. 
In the case of many, high-quality correspondences, classical solvers are typically precise, so the prior has little impact on the solver, while the network learns to heavily rely upon solver output via a low $\wB$.
In the case of few, low-quality correspondences, solvers degrade, so the prior is designed to have a strong influence on solver output, while the network relies more heavily on the transformer predictions (high $\wB$).

The approach is agnostic to input features and correspondences.
In experiments (Sec~\ref{sec:experiments}), we show success with features from three feature estimation methods~\cite{sun2021loftr,rockwell2022,arnold2022map} and correspondences from two correspondence estimators~\cite{sarlin2020superglue,sun2021loftr}.
We use a ViT~\cite{dosovitskiy2020image} to handle spatial features, and losses can backpropagate through the backbone.
We also explore using only correspondences and descriptors as input.

\subsection{Pose Transformer}
\label{sec:approach_learned}

The goal of the Transformer is to estimate (1) 6DoF relative camera pose $\TB_t$ between two wide-baseline images and (2) weight $\wB \in [0,1]$ of its estimate vs. solver estimates from a set of 2D correspondence matches $\mathbbm{M} = \{(\pB,\qB)\} \: | \: \pB,\qB \in \mathbb{R}^2$  and optionally dense 2D image-wise features $f_i, f_j$. 
Given predicted camera pose and weight, the final output is the  linear combination of Transformer pose $\TB_{t}$ and solver pose $\TB_s$ weighted by $\wB$. 
We use separate weights for translation $w_t$ and rotation $w_r$, allowing the Transformer to have different confidences for two subtly different problems.

Two challenges to this setup are that linear combinations of rotations are often not rotation matrices, and solver translation does not have scale.
To address the former, we represent pose in the 6D coordinate system of Zhou \textit{et al.} \cite{zhou2019continuity}, which enables us to combine in 6D space before computing a rotation matrix using Gram--Schmidt orthogonalization. 
To address scale-less solver output, we scale translation $\tB_s$ by the Transformer predicted translation magnitude $||\tB_t||$, before linearly combining. 
We find this stabilizes training compared to first averaging predicted the angles of $\tB_s$ and $\tB_{tf}$ and then applying scale to normalized predictions.
Our final formula is:
\vspace{-3mm}
\begin{equation}
\label{eqn:weighting}
\begin{split}
\hat{\RB} = w_r \RB_t + (1-w_r) \RB_s \\
\hat{\tB} = w_t \tB_t + (1-w_t) ||\tB_t|| \tB_s  \\
\end{split}
\end{equation}
\vspace{-3mm}

\parnobf{Transformer Backbone} 
We use two distinct architectures to span possible inputs: if features are available, we use a modified ViT. 
If only correspondences are available, we use a Vanilla Transformer.
This means the method can be used with correspondence or regression-based methods producing dense features, while accommodating methods only outputting correspondence.
In each case, the Transformer produces features $f_o$ used as input to two MLP heads. 

\parnoit{8-Point ViT} This network takes as input pairwise dense features $f_i, f_j$
and produces a feature vector $f_o$. %\crnote{TODO} $V$. 
It consists first of one LoFTR~\cite{sun2021loftr} self-attention and cross-attention layer followed by an 8-Point ViT cross-attention layer~\cite{rockwell2022,dosovitskiy2020image,lu2019vilbert}; both networks are aimed at producing good features for pose estimation.
For detailed architectures of each, see the original works.

\parnoit{Vanilla Transformer} This network takes as input a set of correspondences $\mathbbm{M} = \{(\pB,\qB)\} \: | \: \pB,\qB \in \mathbb{R}^2$ including associated descriptors, if available, and produces a set of features $f_o$. 
We use a vanilla Transformer encoder with $N$ layers, and map correspondences and descriptors as input.
We encode correspondences in a sinusoidal manner with $K$ bands, followed by a linear mapping to a size of $c$ input to the Transformer.
If descriptive features of each correspondence point are available, of dimension $d < c$, a Linear layer maps them to $\frac{c}{4}$ and they are concatenated to correspondence locations which are linearly mapped to $\frac{3c}{4}$.

The Vanilla Transformer can also be used with dense features $f$ as input. 
If networks produce a joint feature encoding for two images, the Transformer can be applied directly to low-resolution features, without positional encoding.
This occurs in the Regression model of Arnold \textit{et al.}~\cite{arnold2022map}, which we build upon in Table~\ref{tab:mapfree}.

\parnoit{Regression MLP} This MLP maps Transformer features $f_0$ to $\RB \in \mathbb{R}^6$ and $\tB \in \mathbb{R}^3$ using two hidden layers.

\parnoit{Gating MLP} This takes as input Transformer features concatenated with Regression MLP predictions and predictions from the classical solver, along with the number of inlier correspondences in the solver output, using several thresholds.
Predictions and number of inliers are normalized then input as scalar features.
As we analyzed in Figure~\ref{fig:analysis}, the number of inlier correspondences is highly correlated with the performance of solver pose estimate $\TB_s$.
The Gating MLP has two hidden layers and ends with a Sigmoid, producing $w_t, w_r \in (0,1)$.

\subsection{Prior-Guided Robust Pose Estimator}
\label{sec:approach_solver}
Now, having shown how the solver can help learning, how can the learning based methods help the solver?
The performance of search-based solver methods like RANSAC~\cite{Fischler81} is driven by {\it searching} over a model space by sampling valid hypothesis and then ranking them based on a {\it scoring function}. The scoring function serves as the measure of probability of data under a hypothesis~\cite{torr2000mlesac}.
It's typical to use such solvers when estimating pose from a set of correspondences, but direct application of these methods can not be robust when correspondence estimation is done with a scarce set.
Our key idea is to use the predicted pose estimate, $\TB_1$, to influence both the {\it search} and the {\it scoring} function to help in data scarce scenarios. 

We take inspiration from existing lines of work in using learning to better inform sampling and selection in RANSAC-like algorithms~\cite{barath2019magsac, barath2020magsac++, raguram2012usac, brachmann2017dsac, torr2000mlesac}. We show that we can recycle estimates from a learning-based model and plug these estimates in simplistically. Specifically, an initial estimated pose, $\TB_1$, to modify the search function so as to sample more hypothesis close to $\TB_1$. Secondly, we modify the scoring function to consider the distance to the $\TB_1$ along with inlier count.

\parnobf{RANSAC Preliminaries} The typical approach to pose estimation from correspondences applies random sample consensus (and variants) e.g. RANSAC, USAC~\cite{raguram2012usac} or MAGSAC~\cite{barath2019magsac,barath2020magsac++} to model fitting e.g. 5, 7, or 8-Point algorithms~\cite{hartley1997defense,longuet1981computer,nister2004efficient,li2006five}. These methods use a notion of epipolar distance such as Sampson Error~\cite{hartley2003multiple} for inlier thresholding (soft and hard).
More concretely, given a set of 2D correspondence matches $\mathbbm{M} = \{(\pB,\qB)\} \: | \: \pB,\qB \in \mathbb{R}^2$, a minimal subset of points is randomly sampled to fit  a model $\HB$ via an n-point algorithm. The scoring function counts  number of inliers that have Sampson Error less than a fixed threshold $\sigma$. Given, hypothesis $\HB$ and set $\MB$ of correspondences, $\EB(\pB, \qB | \HB)$, is the Sampson Error between points $\pB$ and $\qB$  under  $\HB$. The scoring function is defined as $\textrm{score}(\HB) = \sum_{\{\pB, \qB\} \in \mathbbm{M}}\mathbbm{1}(\EB(\pB,\qB|\HB) < \sigma)$. Sampling repeats up to $N$ times or until stopping heuristics are met for efficiency~\cite{raguram2012usac}, and the highest scoring model, e.g. the one with the most inliers, is selected. Works like MAGSAC~\cite{barath2019magsac}, MAGSAC++~\cite{barath2020magsac++} have shown that improving scoring functions to show better performance. For simplicity, we continue the exposition with thresholding based function that are popular with classic RANSAC~\cite{Fischler81}.

\parnobf{Limitations in Few-Correspondence Case} The heuristic score of counting inliers typically is not effective especially in the low-correspondence case~\cite{barroso2023two}. 
When the number of correspondences is only a small multiple of the number of points needed to minimally define a model the algorithm becomes  particularly unreliable. Consider the extreme case of doing pose recovery with calibrated cameras from nine points, of which five are inliers. The minimal subset for pose estimation is five points, and so while the true model will have five inliers, so will any other sampled hypothesis. Accordingly, the result will be random hypothesis.

\parnobf{Prior-Guided Estimator} We propose to incorporate a learning based predictions to aid the solver in the case of few or poor correspondences. We operationalize this by incorporating a prior model that estimates the likelihood of hypothesis under the network's prediction using a function $\beta(\cdot|\TB_{1})$.
The $\beta(\HB|\TB_{1})$ measures the log probability of the hypothesized model $\HB$ under $\TB_1$. We found it difficult to measure probabilities in rotation and translation and weigh them, so as a proxy, we compare how the models transform a fixed set of grid points. In particular, we measure the negative of average squared distance between a fixed set of grid points transformed by $\TB_1$ and the same fixed grid transformed by $\HB$. See Supplemental for details.

Now, we show how this $\beta$ function can alter the scoring. The modified scoring function measures the likelihood of the hypothesis $\HB$ under $\TB_{1}$ along with measuring the likelihood of data~\cite{torr2000mlesac} under $\HB$. It is defined as,
\begin{equation}
\label{eqn:prior}
\textrm{score}(\HB) = 
\alpha  \beta(\HB | \TB_1) + \hspace{-0.75em}
\sum_{(\pB, \qB)\in \mathbbm{M}}  \hspace{-0.75em} \mathbbm{1}\Bigl(\EB(\pB,\qB|\HB) < \sigma\Bigr) 
\end{equation}
which is the (log) product of probability of the hypothesis given our $\beta$ prior function and the probability of the the data, $\mathbbm{M}$, under $\HB$. We weigh the prior with a scalar, $\alpha \in \mathbb{R}$.  In this setting, the prior tie-breaks ambiguous cases where two hypotheses have similar numbers of inliers, but has diminishing influence as $|\mathbbm{M}|$ gets bigger. As $|\mathbbm{M}| \to \infty$ , the prior's impact is washed out entirely.  This formulation has the desired impact of significant effect when correspondences are few and unbiased hypotheses are poor, and little impact when correspondences are many.

\parnobf{Sampling Good Hypotheses}
Randomly sampling points and estimating $\HB$ is unlikely to lead to hypothesis consistent with the model $\TB_{1}$. To increase the chance to sampling consistent hypothesis  we want to sample a minimal subset that best agrees with the model. We achieve this by weighing the correspondences by their agreement with the model in turn influencing the  as $w(\pB,\qB)  = \exp(-\text{Sampson}(\pB, \qB| \TB_1)/\tau)$.

In practice, we sample half the hypothesis using biased sampling use uniform sampling for the other half.
This improves sample diversity in the case of many correspondences, in which case unbiased sampling is very effective. 

\subsection{Implementation and Training Details}
\label{sec:approach_implementation}

Across experiments, we use the Adam~\cite{kingma2014adam} optimizer.
The 8-Pt ViT trains for about 300k iterations or 7 days on 10 GTX 1080Ti; Vanilla TF trains for about 600k iterations or 3 days.
We select the checkpoint with the lowest validation mean rotation error, which tends to be marginally more stable than translation error.
We represent rotation in 6D coordinates~\cite{zhou2019continuity} and use L1 loss.
Models are trained stage-wise: first we train the Transformer to estimate pose, then to estimate pose jointly with a vanilla solver, then to estimate jointly with the prior-based solver.
We find this progressive training improves final performance.
We implement in PyTorch~\cite{paszke2019pytorch} Lightning~\cite{Falcon_PyTorch_Lightning_2019}, using TIMM~\cite{rw2019timm} for the ViT.

\parnobf{8-Point ViT} We train 8-Point ViT end-to-end with the feature extraction backbone.
We found including a self and cross-attention LoFTR layer significantly improved learning capacity, while additional layers did not help further.

\begin{figure}[t]
    \centering
    \includegraphics[width=\linewidth]{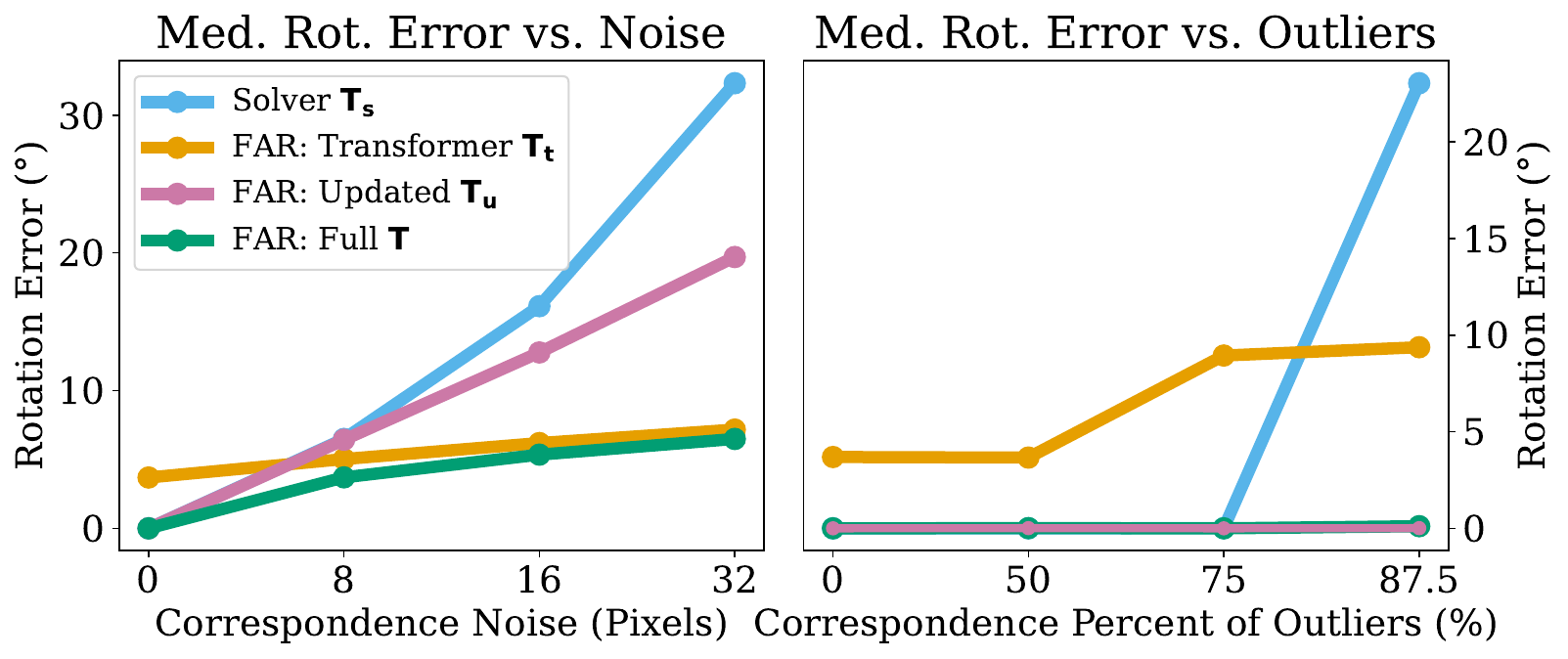}
    \vspace{-1.2em}
    \caption{\textbf{Ground Truth Robustness Study on Matterport3D}. Using true correspondence, the solver is nearly perfect. 
    Adding noise or outliers, it quickly degrades, while prior-guided Updated solver is robust to outliers and the Transformer is robust to noise. 
    FAR matches or beats all methods across settings. 
    }
    \label{fig:theoretical}
    \vspace{-1.2em}
\end{figure}

\parnobf{Vanilla Transformer} We use a 6-layer encoder with 8 heads and 512 feature size followed by global average pooling.
We use $K=42$ bands for positional encoding, and linearly map them to size 384, concatenating descriptive features linearly mapped to size 128.
We found random dropout on correspondences with $p=0.1$ helps performance.
We cache correspondences for fast training. Training speed is ${\approx}$12 iterations per second on a GTX 1080Ti.

\parnobf{Prior-Guided Estimator} We implement in Kornia~\cite{eriba2019kornia} and use 2k random samples without early stopping and inlier threshold on L2 Sampson Error $\sigma$ of $3\times10^{-7}$, finding these results most closely matched OpenCV~\cite{opencv_library} output using LoFTR settings.
For the Prior, we use $\tau=0.1$ and $\alpha=3.33$ found through grid search on the validation set. 
\section{Experiments}
\label{sec:experiments}

\begin{table}
    \caption{\textbf{Camera Pose Estimation on Matterport3D.} Correspondence-based methods have low median but high mean error, and do not produce translation scale. Regression-based methods are less precise but produce scale. FAR builds upon both, resulting in low median and mean error, with translation scale.}
\label{tab:matterport}
\vspace{-0.7em}

\resizebox{\ifdim\width>\columnwidth \columnwidth \else \width \fi}{!}{

\begin{tabular}{l c c c c c c} \toprule
  & \multicolumn{3}{c}{Translation (m)} & \multicolumn{3}{c}{Rotation ($^{\circ}$)} \\
Method & Med.$\downarrow$ & Avg.$\downarrow$ & $\leq$1m$\uparrow$ & Med.$\downarrow$ & Avg.$\downarrow$ & $\leq$ 30$\uparrow$ \\
\midrule
\cite{raposo2013plane} + \cite{ranftl2020towards} & 3.34 & 4.00 & 8.3 & 50.98 & 57.92 & 29.9 \\
Assoc.3D~\cite{qian2020associative3d} & 2.17 & 2.50 & 14.8 & 42.09 & 52.97 & 38.1   \\
Sparse Planes~\cite{jin2021planar} & 0.63 & 1.25 & 66.6 & 7.33 & 22.78 & 83.4   \\
PlaneFormers \cite{agarwala2022planes} & 0.66 & 1.19 & 66.8 & 5.96 & 22.20 & 83.8   \\
8-Point ViT~\cite{rockwell2022} & 0.64 & 1.01 & 67.4 & 8.01 & 19.13 & 85.4 \\ 
NOPE-SAC-Reg~\cite{tan2023nope} & 0.52 & 0.94 & 73.2 & 2.77 & 14.37 & 89.0 \\
SuperGlue~\cite{sarlin2020superglue} & - & - & - & 3.88 & 24.17 & 77.8   \\
LoFTR~\cite{sun2021loftr} & - & - & - & \underline{0.23} & 9.49 & 91.4   \\
LoFTR+Reg. Scale & 0.85 & 1.21 & 56.3 & 0.26 & 9.66 &	91.2 \\
\midrule
FAR (Vanilla TF) & \underline{0.37} & \underline{0.67} & \underline{81.9} & 0.26 & \underline{6.14} & \underline{94.2} \\
FAR & \textbf{0.25} & \textbf{0.49} & \textbf{89.2} & \textbf{0.20} & \textbf{4.93} &	\textbf{95.8}   \\
\bottomrule
\end{tabular}
}
\vspace{-1em}
\end{table}

We design our experiments to measure the effectiveness of FAR in achieving our stated goals: flexible, accurate and robust 6DoF pose estimation.
We first validate robustness by measuring model performance as a function of increasingly perturbed ground truth. 
We next test precision to moderate view change and robustness to large view change by comparing to the state of the art in wide-baseline relative pose. 
Having demonstrated accuracy and robustness, we next verify model flexibility to choice (or lack) of dense feature method, correspondence estimation method, and dataset size.
Finally, we compare to the state of the art on additional indoor and outdoor datasets.

\subsection{Robustness to Correspondence Perturbations}
\label{sec:experiments_theoretical}

Here, we assume the image correspondences are given and study how variations of our method and baselines perform with varying noise-levels applied to the correspondences. 

\parnobf{Dataset} We use image pairs collected using the Habitat~\cite{savva2019habitat} embodied simulator upon Matterport3D~\cite{chang2017matterport3d}, following the setup of Jin \etal~\cite{jin2021planar}. 
It has 32k train / 5k val / 8k test pairs with small to moderate overlap (average 53$^\circ$
rotation, 2.3m translation, 21\% overlap). 
The variety in view change enables the study of both precision upon moderate cases and robustness to highly challenging cases.

\parnobf{Metrics}
Throughout Matterport3D experiments, we report three metrics for rotation and translation: median error, mean error, and percentage of errors within a threshold. 
These are standard metrics, which identify our two qualities of interest: precision (median) and robustness (mean and percentage).
We follow prior work~\cite{jin2021planar,tan2023nope} in using rotation threshold of $30^{\circ}$ and translation threshold of 1m. 
For the ground truth study, for brevity we report median error across a variety of settings, which is an indicative summary of performance. Additional results are in Supplemental.

\parnobf{Setup} Beginning with ground truth correspondences, we  (1) Apply Gaussian noise with standard deviation from 0 to 32 pixels, upon 480x640 images; (2) Replace true correspondences with randomly sampled coordinates (outliers), with P(Outliers) from 0 to 0.875.
Models are trained and evaluated upon each noise and outlier setting independently; e.g. FAR is trained and evaluated four times to make Correspondence Noise Graph in Figure~\ref{fig:theoretical}, left.

\begin{table}
    \caption{\textbf{Ablations on Matterport3D.} (Top) We improve significantly upon LoFTR using a combination of learned and classical. 
    (Middle) This result holds for the case of no input features, where we use the Vanilla TF. 
    (Bottom) Scaling Solver translation is important to FAR performance; selecting separate weights for $R$ and $T$ improves robustness.}
\label{tab:ess_ablations}
\vspace{-0.5em}

\resizebox{\ifdim\width>\columnwidth \columnwidth \else \width \fi}{!}{
\begin{tabular}{l c c c c c c} \toprule
  & \multicolumn{3}{c}{Translation (m)} & \multicolumn{3}{c}{Rotation ($^{\circ}$)} \\
\textit{Transformer: 8-Point ViT} & Med.$\downarrow$ & Avg.$\downarrow$ & $\leq$1m$\uparrow$ & Med.$\downarrow$ & Avg.$\downarrow$ & $\leq$30$\uparrow$ \\
\midrule
LoFTR + Solver + Scale $\TB_s$ & 0.85 & 1.21 & 56.3 & 0.26 & 9.66 &	91.2 \\
FAR: Transformer $\TB_t$ & 0.38 & 0.64 & 85.4 & 4.51 & 9.94 & 94.2 \\
FAR: One Round $\TB_1$ & \textbf{0.25} & \textbf{0.49} & \underline{89.0} & \textbf{0.20} & \underline{5.08} & \underline{95.7} \\
FAR: Updated $\TB_u$ & \textbf{0.25} & 0.50 & 88.4 & \textbf{0.20} & 5.35 & 95.0 \\
FAR: Full $\TB$ & \textbf{0.25} & \textbf{0.49} & \textbf{89.2} & \textbf{0.20} & \textbf{4.93} &	\textbf{95.8}   \\
\\
\textit{Transformer: Vanilla TF}\\
\midrule
LoFTR + Solver + Scale $\TB_s$ & 0.85 & 1.21 & 56.3 & \underline{0.26} & 9.66 & 91.2 \\
FAR: Transformer $\TB_t$ & 0.42 & 0.75 & 79.1 & 3.87 & 10.8 & 92.5 \\
FAR: One Round $\TB_1$ & \textbf{0.37} & \textbf{0.67} & \underline{81.8} & \underline{0.26} & \underline{6.41} & \underline{93.8} \\
FAR: Updated $\TB_u$ & \textbf{0.37} & 0.68 & 81.5 & \textbf{0.25} & 6.69 & 93.7 \\
FAR: Full $\TB$ & \textbf{0.37} & \textbf{0.67} & \textbf{81.9} & \underline{0.26} & \textbf{6.14} & \textbf{94.2} \\
\\
\textit{Prediction Selection} \\
\midrule
Unscaled Solver $t_s$ & 0.31 & 0.55 & 87.4 & 0.21 & 5.09 & \textbf{95.8} \\
One Weight ($w_r=w_t$) & \textbf{0.25} & 0.50 & 88.7 & \textbf{0.20} & 5.04 & \textbf{95.8} \\
FAR & \textbf{0.25} & \textbf{0.49} & \textbf{89.2} & \textbf{0.20} & \textbf{4.93} &	\textbf{95.8}   \\
\bottomrule 
\end{tabular}
}
\vspace{-1.3em}
\end{table}

\parnobf{Ablations} We consider the following cases: 

\parnoit{(1) Solver $\TB_s$} Using LoFTR's solver, i.e., RANSAC~\cite{Fischler81}+5-Point Algorithm~\cite{nister2004efficient}

\parnoit{(2) FAR: Transformer $\TB_t$} Pose Transformer output pose. This is a Vanilla Transformer, since dense features are not available as input; only correspondences.

\parnoit{(3) FAR: Updated $\TB_u$} Solver output using FAR's Prior

\parnoit{(4) FAR: Full $\TB$} Full FAR: learned combin. of $\TB_u$ and $\TB_t$

\parnobf{Ablation Results} Figure~\ref{fig:theoretical} shows Solver has nearly perfect results on ground truth correspondences but is not robust to noise or many outliers.
FAR: Transformer is less precise on ground truth but is more robust as outlier frequency and particularly noise increases.
The prior is highly effective at leaving FAR: Updated robust to outliers, showing close to 0$^{\circ}$ median error even with 87.5\% outliers.
The full method offers the best of all: precise estimation on ground truth correspondences with the best or equal to best robustness to noise and outliers. 

\begin{table}
    \caption{\textbf{Approach Flexibility to Features and Correspondences.} FAR yields improvement using features from 8-Pt ViT or LoFTR; and correspondences from SuperGlue or LoFTR.}
\label{tab:flexible}
\vspace{-0.5em}

\resizebox{\ifdim\width>\linewidth \linewidth \else \width \fi}{!}{

\begin{tabular}{l l l c c c c c c} \toprule
\multirow{2}{*}{Feats.} & \multirow{2}{*}{Corr.} & \multirow{2}{*}{Pose Est.} & \multicolumn{3}{c}{Translation (m)} & \multicolumn{3}{c}{Rotation ($^{\circ}$)} \\
 &  & &  Med.$\downarrow$ & Avg.$\downarrow$ & $\leq$1m$\uparrow$ & Med.$\downarrow$ & Avg.$\downarrow$ & $\leq$ 30$\uparrow$ \\
\midrule
8-Pt ViT & - & 8-Pt ViT & 0.64 & \textbf{1.01} & 67.4 & 8.01 & 19.1 & 85.4 \\ 
8-Pt ViT & SuperGlue & FAR & \textbf{0.62}	& \textbf{1.01} & \underline{68.3} & \textbf{7.02} & \textbf{16.6} & \underline{86.8} \\ 
8-Pt ViT & LoFTR & FAR & \underline{0.63} & \textbf{1.01} & \textbf{68.5} &	\underline{7.06} & \underline{17.0} & \textbf{86.9} \\ 
\midrule
LoFTR & LoFTR & RANSAC+5Pt &  - & - & - & \underline{0.23} & 9.49 & 91.4   \\
- & LoFTR & FAR (Vanilla TF) & \underline{0.37} & \underline{0.67} & \underline{81.9} & 0.26 & \underline{6.14} & \underline{94.2} \\
LoFTR & LoFTR & FAR & \textbf{0.25} & \textbf{0.49} & \textbf{89.2} & \textbf{0.20} & \textbf{4.93} &	\textbf{95.8}   \\

\bottomrule
\end{tabular}
}
\vspace{-0.5em}
\end{table}

\subsection{Wide-Baseline Pose  on Matterport3D}

In this section, we use the same Matterport3D dataset and the metrics used in Sec.~\ref{sec:experiments_theoretical}, but the inputs are images rather than the GT correspondences.

\parnobf{Baselines} We compare against state-of-the-art solver-based and learning-based baselines.
For solver-based methods, we choose the popular LoFTR~\cite{sun2021loftr} and SuperGlue~\cite{sarlin2020superglue}. 
In the learned space, we compare to end-to-end classical-estimation-inspired ViT, 8-Point~\cite{rockwell2022}; planar mapping and optimization methods NOPE-SAC~\cite{tan2023nope}, PlaneFormers~\cite{agarwala2022planes} and Sparse Planes~\cite{jin2021planar}; and 3D reconstruction method Associative3D~\cite{qian2020associative3d}. 
In this set of experiments our FAR builds upon LoFTR backbone and correspondences.
We additionally report results using only correspondence and descriptor as input, as ``FAR (Vanilla TF)".

\parnobf{Results}  Table~\ref{tab:matterport} shows the quantitative results on Matterport3D.
Among the prior works, end-to-end methods such as 8-Point ViT~\cite{rockwell2022} perform well in absolute Translation, while correspondence-solver methods e.g. LoFTR~\cite{sun2021loftr} perform best in rotation. 
FAR sets a new standard in both metrics, surpassing the best prior baseline (NOPE-SAC-Reg) by a large margin. It reduces the median and mean translation errors by about 50\%: from 0.52 to 0.25 and from 0.94 to 0.49, respectively. Additionally, it decreases the mean rotation error by almost 50\% compared to the best prior work (LoFTR), from 9.66 to 4.93. Even with only correspondence available as input, ``FAR (Vanilla TF)" is typically better than all prior work by a large margin.
\begin{figure}[t]
    \centering
    \includegraphics[width=\linewidth]{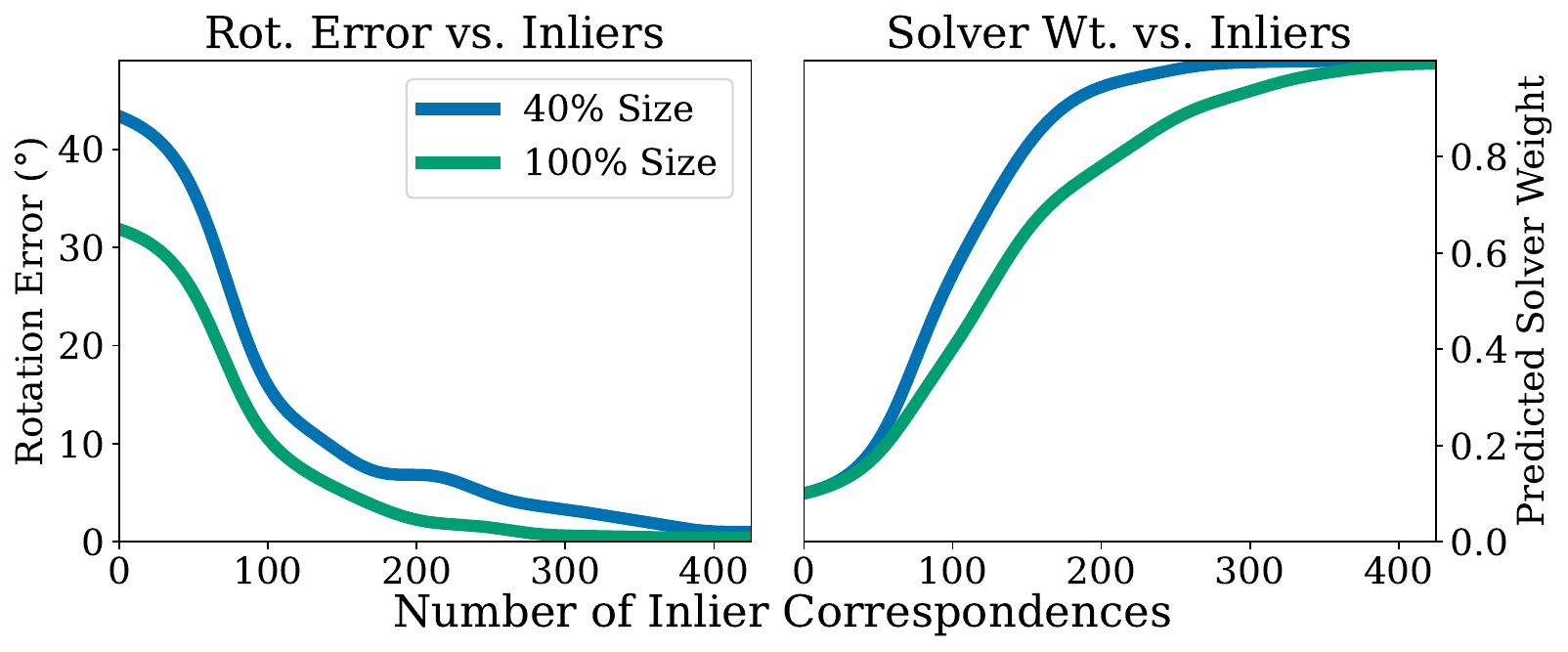}
    \vspace{-1em}
    \caption{\textbf{Evolving with Dataset Size}. 
    The Transformer learns to rely more heavily upon the solver if data is limited (40\% data size), and learns to use Transformer pose estimations as data scales and performance improves (100\% data size).} 
    \label{fig:size}
    \vspace{-1em}
\end{figure}

\parnobf{Ablations} To investigate the source of FAR's out-performance, we conduct ablations on Matterport3D using the same setup of Sec.~\ref{sec:experiments_theoretical}, except correspondences are predicted. 
In addition to ablations discussed in Section \ref{sec:experiments_theoretical}, we consider model output after One Round ($\TB_1$) to study the impact of the prior-guided solver upon the final model.
We further explore performance by using two different Transformer architectures: the dense feature-based Transformer: 8-Point ViT (referred to as FAR in other experiments), and the correspondence-only Transformer: Vanilla TF. 
Finally, we evaluate the design choices outlined in Section \ref{sec:approach_learned} under the Prediction Selection category: Unscaled Solver $t_s$, where Solver translation is not scaled by Transformer magnitude; and One Weight ($w_r=w_t$), which uses equal weights for Transformer translation and rotation prediction.

\parnobf{Ablation Results} As shown in Tab.~\ref{tab:ess_ablations}, we can clearly observe the same trends from Fig.~\ref{fig:theoretical}:
the solver is precise in most cases characterized by low median rotation error. However the solver suffers from high mean error (due to outliers) and poor translation errors. 
Incorporating FAR's prior significantly improves the solver's mean rotation error.
In contrast, Transformer regression outputs are not nearly as precise, with median rotation error of above 4$^\circ$, but it reduces the ratios of large errors (those greater than $1m$ or $30^\circ$).
FAR enhances the best results achieved by both the Transformer and Solver.
These patterns hold true for the Vanilla Transformer as well.
In \textit{Prediction Selection}, we see predicting Solver translation scale is important for translation performance, while separate weighting for rotation and translation improves robustness.

\begin{table*}
    \caption{\textbf{Rotation Performance on InteriorNet and StreetLearn.} Correspondence-based methods (top) struggle to generalize to this data, while regression methods learn helpful features. Building off 8-Point ViT features, we can still utilize LoFTR correspondences to boost performance. Errors were calculated only on successful pairs for SIFT and SuperPoint; gray text indicates failure over 50\% of pairs.}
\label{tab:noah}
\resizebox{\ifdim\width>\linewidth \linewidth \else \width \fi}{!}{
\begin{tabular}{l c c c c c c c c c c c c c c c} \toprule
& \multicolumn{7}{c}{InteriorNet} & & \multicolumn{7}{c}{StreetLearn} \\
& \multicolumn{3}{c}{Large Overlap ($^{\circ}$)} & & \multicolumn{3}{c}{Small Overlap ($^{\circ}$)} & & \multicolumn{3}{c}{Large Overlap ($^{\circ}$)} & & \multicolumn{3}{c}{Small Overlap ($^{\circ}$)} \\
\cmidrule{2-4} \cmidrule{6-8} \cmidrule{10-12} \cmidrule{14-16} 
Method & Med.$\downarrow$ & Avg.$\downarrow$ & $\leq$ 10$\uparrow$ & & Med.$\downarrow$ & Avg.$\downarrow$ & $\leq$ 10$\uparrow$ & & Med.$\downarrow$ & Avg. $\downarrow$ & $\leq$ 10$\uparrow$ & & Med.$\downarrow$ & Avg.$\downarrow$ & $\leq$ 10$\uparrow$ \\
\midrule
SIFT*~\cite{lowe2004distinctive} & 2.95 & 7.78 & 55.5 & & \textcolor{gray}{10.0} & \textcolor{gray}{18.2} & \textcolor{gray}{18.5} & & \textcolor{gray}{3.13} & \textcolor{gray}{18.9} & \textcolor{gray}{22.4} & & \textcolor{gray}{13.8} & \textcolor{gray}{38.8} & \textcolor{gray}{5.7} \\
SuperPoint*~\cite{detone2018superpoint} & 2.79 & 5.46 & 65.9 & & \textcolor{gray}{5.82} & \textcolor{gray}{11.6} & \textcolor{gray}{11.7} & & \textcolor{gray}{1.79} & \textcolor{gray}{6.38} & \textcolor{gray}{16.5} & & \textcolor{gray}{6.85} & \textcolor{gray}{6.80} & \textcolor{gray}{1.0} \\
LoFTR~\cite{sun2021loftr} & \textbf{0.54} & \underline{1.85} & 97.0 & & 2.64 & 14.3 & 70.4 & & 24.8 & 36.4 & 31.6 & & 51.2 & 58.6 & 19.9 \\
Reg6D~\cite{zhou2019continuity} & 6.91 & 10.5 & 67.8 & & 11.4 & 21.9 & 44.1 & & 6.02 & 12.3 & 69.1 & & 7.59 & 15.1 & 63.4 \\
Cai \emph{et al.}~\cite{cai2021extreme} & 1.10 & 2.89 & 97.6 & & \underline{1.38} & 10.2 & 89.8 & & 2.91 & 9.12 & 87.5 & & 3.49 & 13.0 & 84.2 \\
8-Point ViT~\cite{rockwell2022} & 1.83 & 2.90 & \underline{97.9} & & 2.38 & \underline{4.48} & \textbf{96.3} & & \underline{2.43} & \underline{4.08}  & \underline{90.1} & & \underline{3.25} & \underline{9.19} & \underline{87.7} \\
\midrule
FAR & \underline{0.60} & \textbf{1.16} & \textbf{98.5} & & \textbf{1.22} & \textbf{3.42} & \underline{95.4} & & \textbf{1.81} & \textbf{3.01} & \textbf{96.7} & & \textbf{2.07} & \textbf{7.89} & \textbf{92.4} \\
\bottomrule
\end{tabular}
}
\vspace{-1em}
\end{table*}

\begin{figure}[t]
    \centering
    \includegraphics[width=\linewidth]{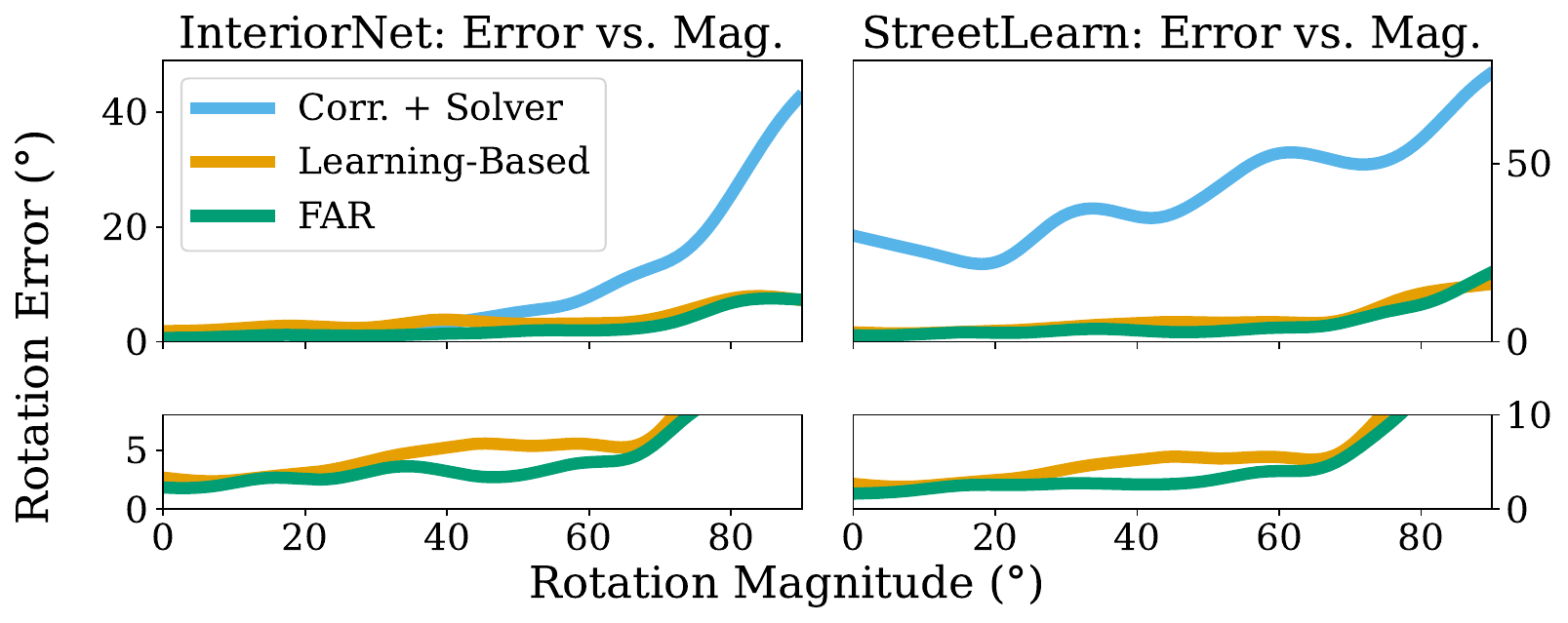}
    \vspace{-1.2em}
    \caption{\textbf{Rotation Error on InteriorNet and StreetLearn}. Even when Correspondence + Solver method is relatively poor, we can still leverage it to improve regression results. ``Learning-Based": 8-Point ViT~\cite{rockwell2022}. ``Corr. + Solver": LoFTR~\cite{sun2021loftr}.}
    \label{fig:interiornet_streetlearn}
    \vspace{-0.6em}
\end{figure}

\subsection{Approach Flexibility}

We then evaluate the flexibility of FAR in terms of the feature extractor, correspondence estimator, and dataset size.

\parnobf{Dataset and Metrics} We continue using the Matterport3D dataset and metrics as in Section~\ref{sec:experiments_theoretical}.

\parnobf{Alternative Approaches} To assess the versatility of FAR, we explore options that are orthogonal to our core contribution. Specifically, we examine three settings for feature estimation: the recent SOTA methods LoFTR and 8-Point ViT, as well as a scenario without dense features. Additionally, we evaluate two recent SOTA settings for correspondence estimation: LoFTR, and SuperPoint~\cite{detone2018superpoint} + SuperGlue~\cite{sarlin2020superglue}.

\parnobf{Results} Table~\ref{tab:flexible} shows FAR improves upon 8-Pt ViT, using either SuperGlue or LoFTR correspondences.
Similarly, FAR improves LoFTR, whether it employs both LoFTR features and correspondences or just the correspondences.

\parnobf{Dataset Scaling} We next present the proposed method when trained on a version of the Matterport3D dataset that has been randomly subsampled to 40\% of its original  data size. 
In Figure~\ref{fig:size}, we compare rotation error (left) and solver weight (right) of 40\% size and full size.
Note that as the training dataset size increases from 40\% to 100\%, both the solver weight and error decrease.
This trend aligns with expectations: as the Transformer's estimated pose accuracy improves with more training data, it gains a larger influence in the final output, enhancing overall performance.
The result suggests a fixed weighting of learned and solver output is not sufficient for best results.

\subsection{Wide-Baseline Pose on Additional Datasets}

We evaluate our method's performance on various datasets to assess its versatility. We follow Cai \etal~\cite{cai2021extreme} and use InteriorNet~\cite{li2018interiornet}, a synthetic collection of indoor home scenes, and StreetLearn~\cite{mirowski2019streetlearn}, which features outdoor city street photos. Both datasets consist of 90$^\circ$ field-of-view image crops upon panoramas. Image pairs are chosen from different panoramas with varying overlaps, facilitating the assessment of precision and robustness in scenarios with both large ($>45^\circ$) and small ($<45^\circ$) overlaps.
Additionally, we use Map-free Relocalization ~\cite{arnold2022map}, a challenging dataset of user-collected videos surrounding outdoor places of interest e.g. sculptures or fountains. SfM has been applied to the videos, so translation with scale can be evaluated.

\begin{table}
    \caption{\textbf{6DoF Performance on Map-free Relocalization}. We compare against the strongest baselines from~\cite{arnold2022map}; for all comparisons see the original paper. FAR uses the 6D Reg backbone, adding LoFTR or SuperGlue correspondences to beat 6D Reg.}
\label{tab:mapfree}

\resizebox{\ifdim\width>\columnwidth \columnwidth \else \width \fi}{!}{

\begin{tabular}{l c c c c c c c c} \toprule

\multirow{2}{*}{Method} & \multicolumn{4}{c}{Pose Error} & & \multicolumn{3}{c}{VCRE} \\
\cmidrule{2-5} \cmidrule{7-9} 
& \multicolumn{2}{c}{Avg. Med.$\downarrow$} & Prec.$\uparrow$ & AUC$\uparrow$ & & Avg. Med.$\downarrow$ & Prec.$\uparrow$ & AUC$\uparrow$ \\
\midrule
LoFTR & 199cm & 30.6$^{\circ}$ & 0.15 & \underline{0.35} & & 168px & 0.35 & 0.61 \\
SuperGlue & 188cm & 25.4$^{\circ}$ & \underline{0.17} & \underline{0.35} & & 160px & 0.36 & 0.60 \\
Reg Ang.~\cite{arnold2022map} & 210cm & 33.7$^{\circ}$ & 0.09 & - & & 200px & 0.30 & - \\
6D Reg~\cite{arnold2022map} & 168cm & 22.5$^{\circ}$ & 0.06 & - & & 147px & 0.40 & - \\
\midrule
FAR (SG) & \underline{149cm} & \textbf{17.2$^{\circ}$} & \underline{0.17} & \underline{0.35} & & \textbf{135px} & \textbf{0.44} & \underline{0.67} \\
FAR (LoFTR)& \textbf{148cm} & \underline{18.1$^{\circ}$} & \textbf{0.18} & \textbf{0.39} & & \underline{137px} & \textbf{0.44} & \textbf{0.68} \\
\bottomrule
\end{tabular}
}
\vspace{-1em}
\end{table}

\begin{figure}
    \centering
    \includegraphics[width=\linewidth]{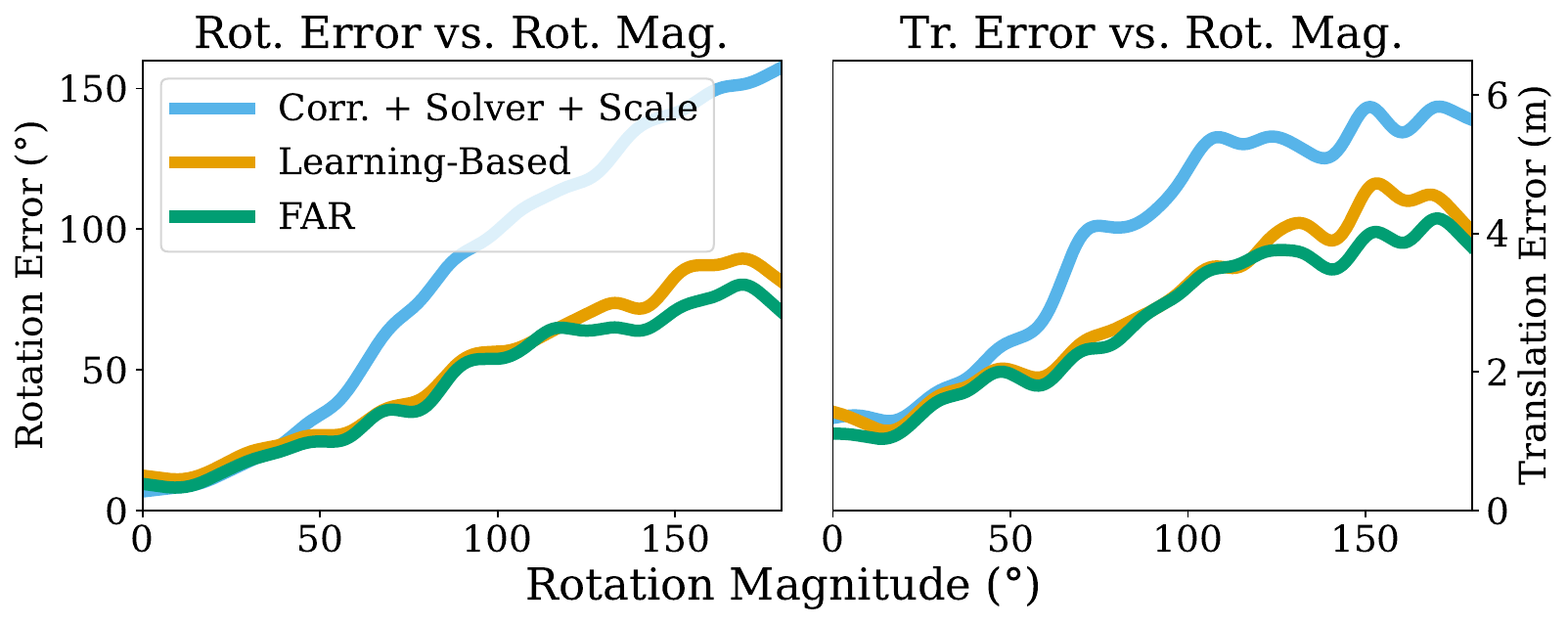}
    \vspace{-1em}
    \caption{\textbf{Error on Map-free Relocalization}. FAR leverages Solver output to improve regression results on this highly challenging dataset. ``Corr. + Solver + Scale": LoFTR + DPT~\cite{ranftl2021vision} trained on KITTI~\cite{geiger2012we}. ``Learning-Based": 6D Reg.}
    \label{fig:mapfree}
    \vspace{-1em}
\end{figure}

\parnobf{Metrics} 
For InteriorNet and StreetLearn, we report rotation error only, in line with prior work~\cite{cai2021extreme,rockwell2022}, using a $10^\circ$ threshold. For Map-free Relocalization, we calculate median translation and rotation errors per video, then average these. We also include the Virtual Correspondence Reprojection Error (VCRE) metric to measure reprojection errors (see~\cite{arnold2022map} for details).

\parnobf{Baselines} 
We compare our method with SOTA correspondence and pose estimation techniques. For InteriorNet and StreetLearn, we compare to Cai \textit{et al.}'s~\cite{cai2021extreme} correlation volume-based learning, SuperPoint~\cite{detone2018superpoint}, and the classical SIFT method~\cite{lowe2004distinctive}. We use LoFTR adapted for InteriorNet using Matterport3D, and for StreetLearn using MegaDepth, due to the lack of depth data in these datasets for training correspondences.
Since LoFTR cannot be finetuned, we find 8-Point ViT features are more effective, and use these as input to FAR, along with LoFTR correspondences.
See Supplemental for details.

Arnold \textit{et al.}~\cite{arnold2022map} train a variety of pose estimation methods on Map-free, including ``6D Reg", which creates a correlation volume~\cite{kendall2017end,khamis2018stereonet} and warps accordingly, followed by a ResNet~\cite{he2016deep}, and is supervised upon 6D rotation~\cite{zhou2019continuity}.

\parnobf{Results} Tab.~\ref{tab:noah} and Fig.~\ref{fig:interiornet_streetlearn} show that 8-Point ViT~\cite{rockwell2022} achieves impressive mean errors, under $5^\circ$, on InteriorNet, even for small overlap pairs.
FAR still adds precision on top of the 8-Point ViT.
On the challenging StreetLearn data, FAR significantly outperforms the state of the art, despite LoFTR not generalizing well to StreetLearn.

\begin{figure}[t]
    \centering
    \includegraphics[width=\linewidth]{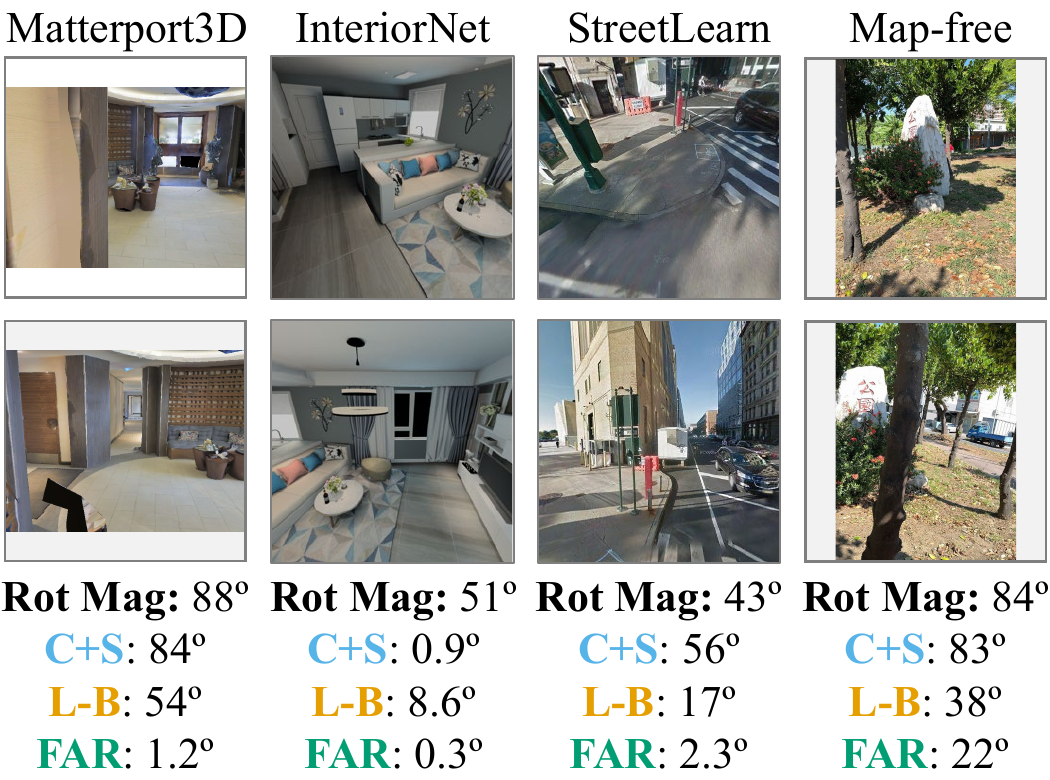}
    \vspace{-1.3em}
    \caption{\textbf{Success Cases}. For some challenging wide-baseline image pairs, our method often dramatically outperforms the baselines.
    ``Learning-Based": LoFTR~\cite{sun2021loftr} with 8-Pt. ViT~\cite{rockwell2022} head (Matterport3D), 8-Pt. ViT  (InteriorNet, StreetLearn), 6D Reg~\cite{arnold2022map}  (Map-free). ``Corr. + Solver": LoFTR.
    }
    \label{fig:datasets}
    \vspace{-1.5em}
\end{figure}

6D Reg is the strongest overall baseline on Map-free Relocalization, so we use its features for FAR, taking correspondences from LoFTR or SuperGlue.
In both cases, FAR improves upon 6D Reg and other methods (see Tab. \ref{tab:mapfree} and Fig. \ref{fig:mapfree}). These results across datasets show FAR's adaptability to different backbones and its robustness to sub-optimal correspondence estimates, highlighted in Fig. \ref{fig:datasets}.

\vspace{-0.5em}
\section{Conclusion}
\label{sec:conclusion}

In this work, we address the problem of 6DoF relative camera pose estimation given a wide-baseline image pair. 
Our introduced FAR represents a simple yet potent approach that merges the best aspects of correspondence-based and learning-based methods. This results in precise and robust outcomes, adaptable to various backbones and solvers.

\parnobf{Limitations and Societal Impact} 
FAR consists of several components and implements Prior-Guided RANSAC in Kornia, slowing inference speed to 3.3 it/sec on 10 1080Ti GPUs; analysis vs. other methods appears in Supplemental Figure~\ref{fig:supp_efficiency}.
Training upon affluent homes of Matterport3D can result in worse performance on more typical residences. 

\parnobf{Acknowledgments and Disclosure of Funding}
Thanks to Jeongsoo Park and Sangwoo Mo for their helpful feedback.
Thanks to Laura Fink and UM DCO for their tireless computing support. 
Toyota Research Institute provided funds to support this work.
{
    \small
    \bibliographystyle{ieeenat_fullname}
    \bibliography{main}
}

\clearpage
\begin{appendices}
    This Supplement includes additional detail for the method and experiments, as well as additional experiments and explanation too long for the main paper.
    
    \section{Network Architecture}

We detail the full model along with ablations below as a function of their components; these correspond to predicted pose boxes in Figure~\ref{fig:method_overview}.
Architecture is overviewed in Tables~\ref{tab:model_arch}-\ref{tab:model_arch_farvan} and detailed in Tables~\ref{tab:model_arch_8pt}-\ref{tab:model_arch_scale}.

\parnobf{Solver $\TB_s$} 
Pose estimation from a correspondence estimator followed by a Kornia~\cite{eriba2019kornia} implementation of RANSAC + 5-Point Algorithm, optionally scaled by predicted translation scale. 
Compare to FAR: Full $\TB$, this ablation does not use the Transformer, nor combine Solver and Transformer predictions, nor do a second round of prior-guided solver and combining with Transformer.

We refer to this as ``Solver" if it uses perturbed ground truth correspondences as input (Figure~\ref{fig:theoretical}), meaning no correspondence estimator is used.
We refer to this as Corr. + Solver in experiments if correspondence estimator is used (Figure~\ref{fig:interiornet_streetlearn} and ~\ref{fig:datasets}).
We refer to it as Corr. + Solver + Scale if predicted scale is used to evaluate absolute translation error (Figures~\ref{fig:teaser},~\ref{fig:analysis},~\ref{fig:mapfree}; Table~\ref{tab:ess_ablations}); we refer to it as LoFTR + Solver + Scale if LoFTR is used (Table~\ref{tab:ess_ablations}). 

Predicted scale for Solver $\TB_s$ is the output of the Translation Scale Predictor network detailed in Table~\ref{tab:model_arch_scale}.
It takes dense features $f$ as input and outputs a single scalar, which is multiplied by translation angle output from RANSAC + 5-Point to obtain final translation.
Early in experiments, we used a transformer-based architecture to predict scale, but found this CNN-based method performed better. 

\parnobf{FAR: Transformer $\TB_t$} 
Predicted 6DoF pose from FAR's Transformer.
Compare to FAR: Full $\TB$, this ablation does not use the Solver, nor combine Solver and Transformer predictions, nor do a second round of prior-guided solver and combining with Transformer.

In the case dense features are available, the 8-Point ViT is used, if only correspondences plus descriptors are available, the Vanilla TF is used.
Each is detailed in Table~\ref{tab:model_arch_8pt} and Table~\ref{tab:model_arch_van}, respectively.

\parnobf{FAR: One Round $\TB_1$} 
Predicted 6DoF pose from one round of FAR, which consists of the weighted linear combination in 6D~\cite{zhou2019continuity} space of $\TB_t$ and $\TB_s$, weighted by $w$ as described in Equation~\ref{eqn:weighting}.
Compare to FAR: Full $\TB$, this ablation does not do a second round of prior-guided solver and combining with Transformer.

\parnobf{FAR: Updated $\TB_u$} 
Pose estimation from FAR's prior-guided RANSAC + 5-Point Algorithm, using $\TB_1$ as a prior.
Compare to FAR: Full $\TB$, this ablation does not do a second round of combining with the Transformer.
Note: results from FAR: Updated $\TB_u$ tend to be less accurate than FAR: One Round $\TB_1$.
This is expected, as FAR: Updated $\TB_u$ is intended to be used in combination with Transformer output $\TB_t$ to form final output.
In other words, our goal of FAR: Updated $\TB_u$ is to improve upon Solver $\TB_s$, resulting in better final output after combining with $\TB_t$.

\parnobf{FAR: Full $\TB$} Final predicted 6DoF pose consisting of the weighted linear combination of $\TB_t$ and $\TB_u$, weighted by $w$.

For LoFTR Feature Extractor and Correspondence Estimator, we use $H=480,W=640$ and $D=256,h=60,w=80$, except on Map-free Relocalization experiments, where images are size $H=720,W=544$, so using the same downsampling, $h=90,w=68$.
For SuperGlue Correspondence Estimator, we use $H=480,W=640$.
For 8-Point ViT Feature Extractor (InteriorNet and StreetLearn experiments), we use $H=224,W=224$ and $D=192,h=24,w=24$.
For 6D Reg Feature Extractor (Map-free Relocalization experiments), we use $H=360,W=270$ and $D=256,h=12,w=9$.
Map-free Relocalization setup differs slightly from other setups to use 6D Reg features as input rather than LoFTR or 8-Point ViT, and 6D Reg produces a single feature vector for a pair of images rather than two; for details see Section~\ref{additional}.
In Table~\ref{tab:model_arch_8pt}, we break down the architecture of Transformer $\TB_t$. 8-Point ViT output has $d=D/n_h+p_e=70$, where $n_h=3$ is the number of heads, and $p_e=6$ is the size of positional encodings.

\begin{table*}[t]
\caption{\textbf{Model Architecture: FAR}. 
High-level first defined, followed by detailed components. 
$N$ varies depending on the number of correspondences.
For LoFTR Feature Extractor and Correspondence Estimator, we use $H=480,W=640,D=256,h=60,w=80$. 
Variables for alternative experiments described in text.
}\label{tab:model_arch}
\centering
\begin{tabular}{lccc}
\toprule
\multicolumn{4}{c}{Overview} \\
 Operation & Inputs & Outputs & Output Shape           \\ \midrule
 Input Image & - & - & $ 2 \times 3 \times H \times W$ \\
 Feature Extractor & Input Image & $f_i, f_j$ & $ 2 \times D \times h \times w $ \\
 Correspondence Estimator & Input Image & $\mathbbm{M}$ & $N \times 4$ \\
 8-Point ViT $\TB_t$ & $f_i, f_j$ & $\TB_t$, $w$ & $9$, $2$ \\
 Solver $\TB_s$ & $\mathbbm{M}$ & $\TB_s$ & $9$ \\
 One Round $\TB_1$ & $\TB_s$, $\TB_t$, $w$ & $\TB_1$ & $9$ \\
 Updated $\TB_u$ & $\mathbbm{M}$, $\TB_1$ & $\TB_u$ & $9$ \\
 Full $\TB$ & $\TB_u$, $\TB_t$, $w$ & $\TB$ & $9$ \\
\bottomrule
\end{tabular}

\vspace{3mm}
\caption{\textbf{Model Architecture: FAR (Vanilla TF)}.
}
\vspace{-3mm}
\label{tab:model_arch_farvan}
\begin{tabular}{lccc}
\toprule
\multicolumn{4}{c}{Overview} \\
 Operation & Inputs & Outputs & Output Shape           \\ \midrule
 Input Image & - & - & $ 2 \times 3 \times H \times W$ \\
 Correspondence Estimator & Input Image & $\mathbbm{M}$ & $N \times 4$ \\
 Vanilla Transformer $\TB_t$ & $\mathbbm{M}$ & $\TB_t$, $w$ & $9$, $2$ \\
 Solver $\TB_s$ & $\mathbbm{M}$ & $\TB_s$ & $9$ \\
 One Round $\TB_1$ & $\TB_s$, $\TB_t$, $w$ & $\TB_1$ & $9$ \\
 Updated $\TB_u$ & $\mathbbm{M}$, $\TB_1$ & $\TB_u$ & $9$ \\
 Full $\TB$ & $\TB_u$, $\TB_t$, $w$ & $\TB$ & $9$ \\
\bottomrule
\end{tabular}

\vspace{3mm}
\caption{\textbf{Model Architecture: Transformer $\TB_t$ (8-Point ViT)}. 
}\label{tab:model_arch_8pt}
\vspace{-3mm}
\begin{tabular}{lccc}
\toprule
\multicolumn{2}{c}{Overview} \\
 Operation & Inputs & Outputs & Output Shape           \\ \midrule
 Input Features & - & $f_i, f_j$ & $ 2 \times D \times h \times w $ \\
 LoFTR~\cite{sun2021loftr} Self-Attn. Block & $f_i, f_j$ & $f_{i,1}, f_{j,1}$ & $ 2 \times D \times h \times w $ \\
 LoFTR Cross-Attn. Block & $f_{i,1}, f_{j,1}$ & $f_{i,2}, f_{j,2}$ & $ 2 \times D \times h \times w $ \\
 8-Point ViT~\cite{rockwell2022} Cross-Attn. Block & $f_{i,2}, f_{j,2}$ & $f_o$ & $2 \times D \times d $ \\
 Regression MLP & $f_o$ & $\TB_t$ & $9$ \\
 Gating MLP & $f_o$ & $w$ & $2$ \\
\bottomrule
\end{tabular}

\vspace{3mm}
\caption{\textbf{Model Architecture: Transformer $\TB_t$ (Vanilla TF)} Correspondences optionally include descriptors. If do not, skip Linear Layer, use only Positional Encoding as input to Vanilla Transformer. 
}\label{tab:model_arch_van}
\vspace{-3mm}
\begin{tabular}{lccc}
\toprule
\multicolumn{2}{c}{Overview} \\
 Operation & Inputs & Outputs & Output Shape           \\ \midrule
 Input Corr. & - & $\mathbbm{M}$ & $ N \times 2 \times 2 $ \\
 Input Descriptor & - & $\mathbbm{M}_d$ & $ N \times 2 \times 256 $ \\
 Positional Encoding & $\mathbbm{M}$ & $f_{pos}$ & $N \times 384$ \\
 Linear Layer & $\mathbbm{M}_d$ & $f_{in}$ & $ N \times 128 $ \\
 Vanilla Transformer & $f_{pos}$, $f_{in}$ & $f_{tmp}$ & $ N \times 512$ \\
 Global Avg. Pooling & $f_{tmp}$ & $f_o$ & $ 512$ \\
 Regression MLP & $f_o$ & $\TB_t$ & $9$ \\
 Gating MLP & $f_o$ & $w$ & $2$ \\
\bottomrule
\end{tabular}
\end{table*}

\begin{table*}[t]
\centering

\vspace{3mm}
\caption{\textbf{Model Architecture: Regression MLP}. 
}\label{tab:model_arch_regmlp}
\vspace{-3mm}
\begin{tabular}{lccc}
\toprule
\multicolumn{2}{c}{Overview} \\
 Operation & Inputs & Outputs & Output Shape \\ \midrule
 Input Features & - & $f_o$ & $shape(f_o)$ \\
 Linear & $f_o$ & $f_{tmp0}$ & $512$ \\
 ReLU & $f_{tmp0}$ & $f_{tmp1}$ & $512$ \\
 Linear & $f_{tmp1}$ & $f_{tmp2}$ & $512$ \\
 Linear & $f_{tmp2}$ & $f_{tmp3}$ & $512$ \\
 ReLU & $f_{tmp3}$ & $f_{tmp4}$ & $512$ \\
 Linear & $f_{tmp4}$ & $\TB_t$ & $9$ \\
\bottomrule
\end{tabular}

\vspace{3mm}
\caption{\textbf{Model Architecture: Gating MLP} Shape of $f_o$ is 512 in the case of Vanilla Transformer and $D $ in the case of 8-Point ViT. 
}\label{tab:model_arch_gatmlp}
\vspace{-3mm}
\begin{tabular}{lccc}
\toprule
\multicolumn{2}{c}{Overview} \\
 Operation & Inputs & Outputs & Output Shape \\ \midrule
 Input Features & - & $f_o$ & $shape(f_o)$ \\
 Input Transformer Predicted Pose $\TB_t$ & - & $\TB_t$ & $9$ \\
 Input Solver Predicted Pose $\TB_s$ & - & $\TB_s$ & $9$ \\
 Input Number of Solver Inliers & - & $n_i$ & $3$ \\
 Linear & $f_o$, $\TB_t$, $\TB_s$, $n_i$ & $f_{tmp0}$ & $512$ \\
 ReLU & $f_{tmp0}$ & $f_{tmp1}$ & $512$ \\
 Linear & $f_{tmp1}$ & $f_{tmp2}$ & $512$ \\
 ReLU & $f_{tmp2}$ & $f_{tmp3}$ & $512$ \\
 Linear & $f_{tmp3}$ & $w_{tmp}$ & $2$ \\
 Sigmoid & $w_{tmp}$ & $w$ & $2$ \\
\bottomrule
\end{tabular}

\vspace{3mm}
\caption{\textbf{Model Architecture: Scale Prediction Network}. 
}\label{tab:model_arch_scale}
\vspace{-3mm}
\begin{tabular}{lccc}
\toprule
\multicolumn{2}{c}{Overview} \\
 Operation & Inputs & Outputs & Output Shape           \\ \midrule
 Feature Extractor Input & - & $f_i, f_j$ & $ 2 \times D \times h \times w $ \\
 MaxPool2D & $f_i, f_j$ & $f_{i,a}, f_{j,a}$ & $2 \times D \times h/2 \times w/2$ \\
 Conv2D & $f_{i,a}, f_{j,a}$ & $f_{i,b}, f_{j,b}$ & $2 \times D/2 \times h/2 \times w/2$ \\
 ReLU & $f_{i,b}, f_{j,b}$ & $f_{i,c}, f_{j,c}$ & $2 \times D/2 \times h/2 \times w/2$ \\
 MaxPool2D & $f_{i,b}, f_{j,b}$ & $f_{i,c}, f_{j,c}$ & $2 \times D/2 \times h/4 \times w/4$ \\
 Conv2D & $f_{i,c}, f_{j,c}$ & $f_{i,d}, f_{j,d}$ & $2 \times D/4 \times h/4 \times w/4$ \\
 ReLU & $f_{i,d}, f_{j,d}$ & $f_{i,e}, f_{j,e}$ & $2 \times D/4 \times h/4 \times w/4$ \\
 Conv2D & $f_{i,e}, f_{j,e}$ & $f_{i,f}, f_{j,f}$ & $2 \times D/16 \times h/16 \times w/16$ \\
 ReLU & $f_{i,f}, f_{j,f}$ & $f_{i,g}, f_{j,g}$ & $2 \times D/4 \times h/16 \times w/16$ \\
 Linear & $f_{i,g}, f_{j,g}$ & $f_0$ & 512 \\
 ReLU & $f_0$ & $f_1$ & 512 \\
 Linear & $f_1$ & $f_2$ & 512 \\
 ReLU & $f_2$ & $f_3$ & 512 \\
 Linear & $f_3$ & $s$ & $1$ \\
\bottomrule
\end{tabular}

\end{table*}

\section{Prior-Guided Robust Pose Estimator}
In our implementation of prior guided pose estimation we use RANSAC as the solver to search over the hypothesis space and also score our models with inlier scores. 
We use the five-point algorithm to estimate the Essential Matrix~\cite{hartley1992estimation}. 
Choosing the five-point algorithm is beneficial in the case of known intrinsics (available in all datasets we use) as it only requires 5 correspondences to estimate a minimal model. This increases the chance of sampling a better hypothesis $\HB$ over multiple RANSAC iterations. 
The five-point algorithm recovers the essential matrix corresponding to a minimal set (5) and we convert this to a translation and rotation matrix (up to scale).

\parnobf{Prior Probability} The $\beta(\HB|\TB_{1})$ measures the log probability of the hypothesized model $\HB$ under $\TB_1$. The $\HB$ is the essential matrix and $\TB_{1}$ is the 6D transformation matrix from round of prediction. Since it is difficult to measure the probability of $\HB$ under $\TB_{1}$ we design a proxy formulation. We simplify the formulation with by computing the implied transforms $\{\TB_{\{\HB, k\}}\}_{k=1}^{2}$ corresponding to each of two possible solutions for the rotation matrix.

There are multiple possible ways to measure the probability of the transform $\TB_{\HB, k}$  under $\TB$, one possible solution is to independently measure the distribution for rotation and translation component. This approach however requires tuning different weights for each of the components. In our case we measure the difference in the two transformation by computing the implied effect of the transformations on a given point set.

Specifically, for a randomly sampled  point set $\GB \equiv \{\gB\}_{l=1}^{L}$ in $\mathbb{R}^{3}$ such that $\gB_{l} \in (-3,-3)^{3}$. We transform these points with $\TB_{\{\HB, k\}}$ and $\TB_{1}$. We then compute the squared residuals, ${r}_{i}^{N}$, as distance between the transformed point sets. Assuming the distribution of residuals to be proxy for the pose prior, we can now compute the probability of  residuals under a standard Gaussian distribution as,
\begin{equation}
\beta'(\TB_{\HB, k}, \TB) = \log(\prod_{l=1}^{L}\exp(-r_{l}^{2}) / Z),
\end{equation}
$Z$ is the normalization constant for the probability distribution. We have two hypothesis corresponding to each $\HB$ so we choose solution with the highest log likelihood that best fits with prior to recover $\beta(\HB, \TB)$ as,
\begin{equation} 
\beta(\HB, \TB) = \max \Biggl( \beta'(\TB_{\{\HB, 1\}}, \TB), \beta'(\TB_{\{\HB, 2\}}, \TB)\Biggr)
\end{equation}

\parnobf{Scoring Function} Using the prior score above now we can combine this with our existing RANSAC inliers scoring function by combining the log likelihood for the hypothesis $\HB$ under $\TB$ and the likelihood of correspondence set $M$ under the hypothesis $\HB$ as,
\begin{equation} 
    \text{score}(\HB) =\alpha \beta(\HB, \TB) + \sum_{(\pB, \qB)\in \mathbbm{M}}  \hspace{-0.75em} \mathbbm{1}\Bigl(\EB(\pB,\qB|\HB) < \sigma\Bigr),
\end{equation}
here $\sigma$ denotes the inlier threshold and $\EB(\pB,\qB|\HB)$ measures the Sampson error for correspondences $\pB,\qB$ under the essential matrix $\HB$
    \section{Additional Experimental Details}
\label{addtl}

Our typical training procedure is to train the Correspondence Estimator, followed by FAR: Transformer $\TB_t$ jointly with the backbone, followed by FAR: One Round $\TB_1$, followed by FAR: Full $\TB$.
At each step, we train until validation mean rotation error plateaus, and reload the existing components for the next round of training.
In some cases different steps are not applicable e.g. we build upon learned pose backbone in Map-free Relocalization and cannot train the prior on StreetLearn or InteriorNet.
We use OneCycleLR~\cite{smith2019super} scheduler, except if using 6DReg backbone; here we follow prior work~\cite{arnold2022map} in using a constant learning rate.
FAR's Kornia-based solver is slower than OpenCV, so for speed we use OpenCV solver to compute $\TB_s$ in our final model.
For fair comparison in ablations, we compute $\TB_s$ using Kornia.

\subsection{Ground Truth Robustness Study}
\label{gtrs}

In this experiment, we are given correspondences as input, so we proceed directly to training FAR: Transformer $\TB_t$ and remaining steps. 
Training upon perturbed ground truth correspondences typically plateaus after 90 epochs for FAR: Transformer $\TB_t$ and FAR: One Round $\TB_1$;
we find 10 epochs of additional training is sufficient for FAR: Full $\TB$.
We report $\TB_t$ output after FAR: Transformer $\TB_t$ training rather than after training with the other components in Far: Full $\TB$.
This is because after full training, $\TB_t$ is inaccurate standalone, since it is trained to be effective in conjunction with $\TB_s$.
We report $\TB_u$ and $\TB$ after full training of $\TB$.
We use ground truth correspondence computed via LoFTR's correspondence algorithm from true pose and depth, which consists of a mutual nearest neighbor check.

\subsection{Wide-Baseline Pose Estimation on Matterport3D}

On the full dataset, we found LoFTR reached best performance after 30 epochs, FAR: Transformer $\TB_t$ reached best performance after 39 epochs, FAR: One Round $\TB_1$ reached best performance after 32 epochs, FAR: Full $\TB$ plateaued after 14 epochs.
If using the Vanilla Transformer, FAR: Transformer $\TB_t$ reached best performance after 89 epochs, FAR: One Round $\TB_1$ reached best performance after 69 epochs, FAR: Full $\TB$ plateaued after 12 epochs.
We report $\TB_t$ after its training for the reasons detailed in ~\ref{gtrs}.
In addition, we report $\TB_s$ output after Correspondence Estimator training for consistency with the Correspondence + Solver baseline throughout the paper.
This has little impact upon results compared to reporting after full training of $\TB$.

\subsection{Approach Flexibility}
\label{flex}

\parnobf{Flexibility to Features and Correspondences}
8-Point ViT features refers to spatial features after all self-attention layers in the 8-Point ViT backbone, since the cross-attention layer in 8-Point ViT produces only a flattened array of features.
Given this input, FAR: Transformer $\TB_t$ uses the 8-Point ViT variant.
This normally consists of a LoFTR layer followed by 8-Point ViT cross-attention. 
However, in this special case of 8-Point ViT input, we drop the LoFTR layer to make FAR: Transformer $\TB_t$ equivalent to full 8-Point ViT output.
This allows for closer comparison to the original 8-Point ViT work, while using a specialized architecture, in which inserting a LoFTR layer would not likely be helpful.
FAR: Full $\TB$ can then use FAR: Transformer $\TB_t$ combined with FAR: Updated, with solver output coming from either LoFTR or SuperGlue.

We follow 8-Point ViT training procedure for the model, training for 120k iterations with batch size 60, or about 225 epochs.
We then repeat this procedure for FAR: One Round $\TB_1$ given correspondences from LoFTR or SuperGlue.
Finally, we train for 20k iterations for FAR: Full $\TB$.

\parnobf{Dataset Size}
On the 40\% sized dataset, we found LoFTR reached best performance after 86 epochs, FAR: Transformer $\TB_t$ reached best performance after 94 epochs, FAR: One Round $\TB_1$ reached best performance after 43 epochs, FAR: Full $\TB$ plateaued after 27 epochs.

\subsection{Wide-Baseline Pose Estimation on Additional Datasets}
\label{additional}

\parnobf{InteriorNet and StreetLearn}
We use 8-Point ViT as our feature extractor on InteriorNet and Streetlearn as it is SOTA and correspondence-based methods such as LoFTR cannot be trained on the data as it does not contain depth. 
We follow the training setup of Section~\ref{flex}, training FAR: Transformer $\TB_T$ and FAR: One Round $\TB_1$ sequentially, reloading at each new phase of training, and following 8-Point ViT training schedule for each phase.
We cannot use the prior since it requires translation prediction, which cannot be supervised, since the dataset does not contain translation information.
Therefore, FAR: Full $\TB$ is the output from FAR: One Round $\TB_1$.
However, we find results are strong even without prior.
On InteriorNet, we use LoFTR pretrained on Matterport3D for correspondences.
On StreetLearn, correspondences come from LoFTR pretrained on MegaDepth.

\parnobf{Map-free Relocalization}
We use 6D Reg as our feature extractor for similar reasons to InteriorNet and StreetLearn: 6D Reg has most competitive rotation and translation errors of existing methods, and correspondence-estimation methods such as LoFTR or SuperGlue cannot be trained on the dataset since it does not contain depth.

\begin{figure}[t]
    \centering
    \includegraphics[width=\linewidth]{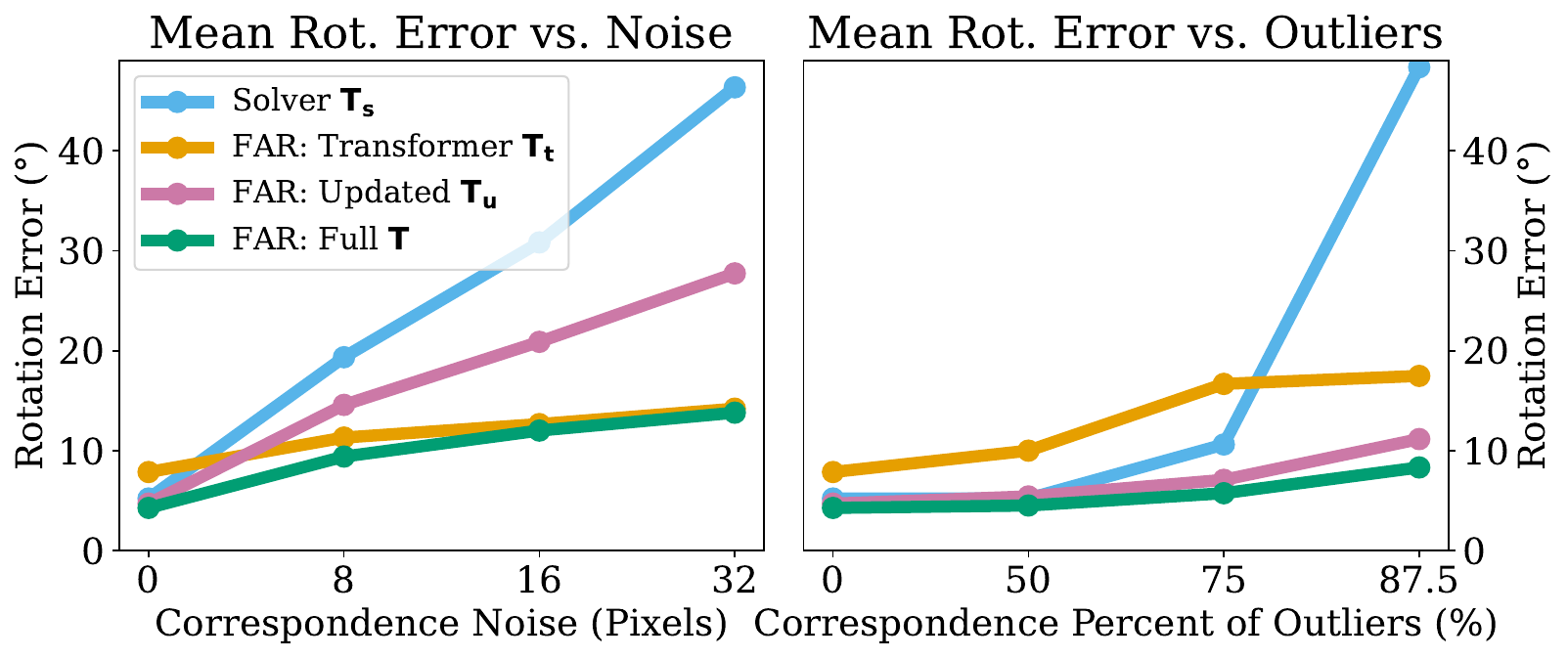}
    \vspace{-1.2em}
    \caption{\textbf{Ground Truth Robustness Study on Matterport3D: Mean Rotation Error}. Using true correspondence, the solver has low mean error, which is nonzero because of some image pairs having limited ground truth correspondences, leading to small mean error. As with median error, adding noise or outliers causes it to quickly degrade, while prior-guided Updated solver is robust to outliers and Transformer is robust to noise. FAR matches or beats all methods across settings. 
    }
    \label{fig:robustness_mean}
    \vspace{-1.2em}
\end{figure}

6D Reg architecture is different from 8-Point ViT and LoFTR in that it warps features to a common frame before estimating pose.
This setup is distinct from the canonical setting of having two dense feature matrices as input, but FAR can nevertheless be adapted.
FAR's Transformer $\TB_t$ takes features after the penultimate ResNet layer of 6D Reg, which yields feature size of $12 \times 9 \times 256$.
The Transformer is a Vanilla Transformer consisting of 6 Transformer Encoder layers with feature size 256 and 8 heads.
We choose the penultimate layer as input to the Transformer as these late features are instructional for predicting pose, and are of feasible resolution and feature size for a Transformer.
The Vanilla Transformer is lightweight, allowing this to be added to a light 6D Reg architecture without significantly impacting run-time or batch size.

We begin from a 6D Reg backbone pretrained for 50 epochs, train FAR: Transformer $\TB_t$ for 20 epochs, train FAR: One Round $\TB_1$ for 30 epochs (50 if using SuperGlue correspondences; which runs faster), followed by another 3 for FAR: Full $\TB$. Correspondences come from either LoFTR or SuperGlue, both of which are pretrained on outdoor environments.
SuperGlue is faster than LoFTR, leading to faster network speed during training and more iterations in the same amount of time. 
FAR: Full $\TB$ training is slower given Kornia solver, so we train for only 3 epochs. 
We nevertheless find this training beneficial.

    \section{Additional Results}

\subsection{Ground-Truth Robustness Experiment} Figure~\ref{fig:robustness_mean} presents mean rotation errors of methods confronted with ground truth correspondences, with varying amounts of noise and outliers. It corresponds to Figure~\ref{fig:theoretical}, except mean rotation error is reported here rather than median rotation error in Figure~\ref{fig:theoretical}. 
The results correspond to those in Figure~\ref{fig:theoretical}: the solver is strong faced with little noise or few outliers, but degrades severely. 
Prior-guided Updated solver is robust to outliers, while Transformer is more robust to noise, at the expense of precision.
FAR produces the best of both results in either low perturbation or high perturbation.
Note solver median errors are 0, but mean errors are nonzero due to image pairs occasionally having very few ground truth correspondences, producing high errors accordingly.
However, this does not impact the experimental conclusion.

\begin{figure}[t]
    \centering
    \includegraphics[width=\linewidth]{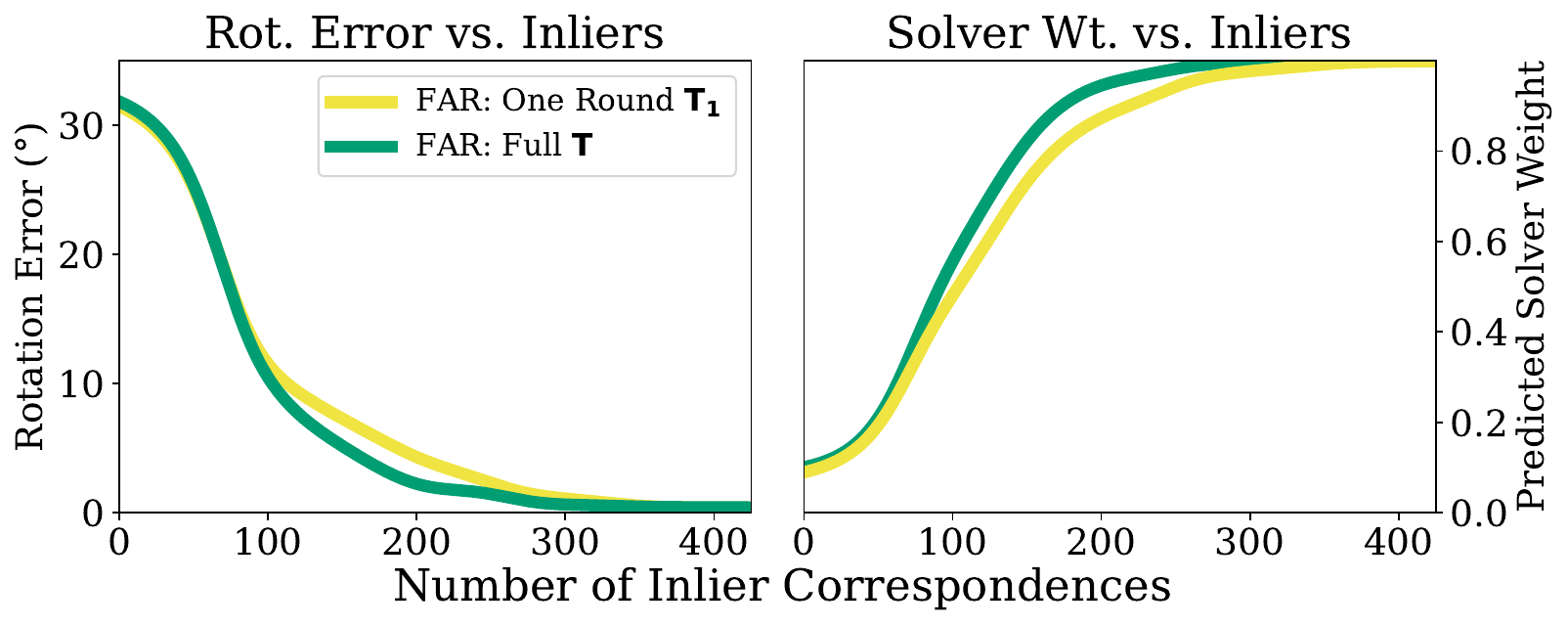}
    \vspace{-1.2em}
    \caption{\textbf{FAR: Full $\TB$ vs. FAR: One Round $\TB_1$}. After one round, FAR produces high-quality results, making further improvement difficult. However, a second forward pass through the solver injected with a prior (FAR: Full $\TB$) improves solver estimates.
    Correspondingly, the Transformer learns to give more weight to the solver (right), and there is a nontrivial improvement in rotation error in difficult cases (left, 100-250 inliers). 
    }
    \label{fig:ours_vs_one_round}
    \vspace{-1.2em}
\end{figure}

\subsection{FAR: Full vs. FAR: One Round} 
Figure~\ref{fig:ours_vs_one_round} displays an analysis of FAR: Full $\TB$ vs. FAR: One Round $\TB_1$. 
The distinction between these baselines is highlighted in Figure~\ref{fig:method_overview}, which is that FAR: Full $\TB$ adds an additional forward pass to the solver, this time injected with the prior. Like FAR: One Round $\TB_1$, this is followed by combination with Transformer predictions.

Despite the two variants of the method being similar, and results of FAR: One Round $\TB_1$ being highly competitive, FAR: Full $\TB$ yields improvement. 
This is apparent in the case of 100-250 inliers, where the prior improves solver output, causing the Transformer to rely on it more (Figure~\ref{fig:ours_vs_one_round}, right)
and the full model to improve (Figure~\ref{fig:ours_vs_one_round}, left).

Note FAR: Full $\TB$ has different weightings $w$ than FAR: One Round $\TB_1$. 
This is because FAR: Full $\TB$ is trained to predict final output given prior-guided solver output.
Since prior-guided solver output is more accurate than vanilla solver output, the network learns to rely upon it more heavily.
For fair comparison with FAR: Full $\TB$, we finetune FAR: One Round $\TB_1$ using a Kornia solver for an equal number of epochs used to finetune FAR: Full $\TB$; before using the Kornia solver during inference.
This is necessary because, as detailed in ~\ref{addtl}, FAR: Full $\TB$ uses cv2's unbiased solver during the first round of computation for efficiency.

\subsection{Inference Time Efficiency Comparison} 

\begin{figure}
\begin{center}
\includegraphics[width=\columnwidth]{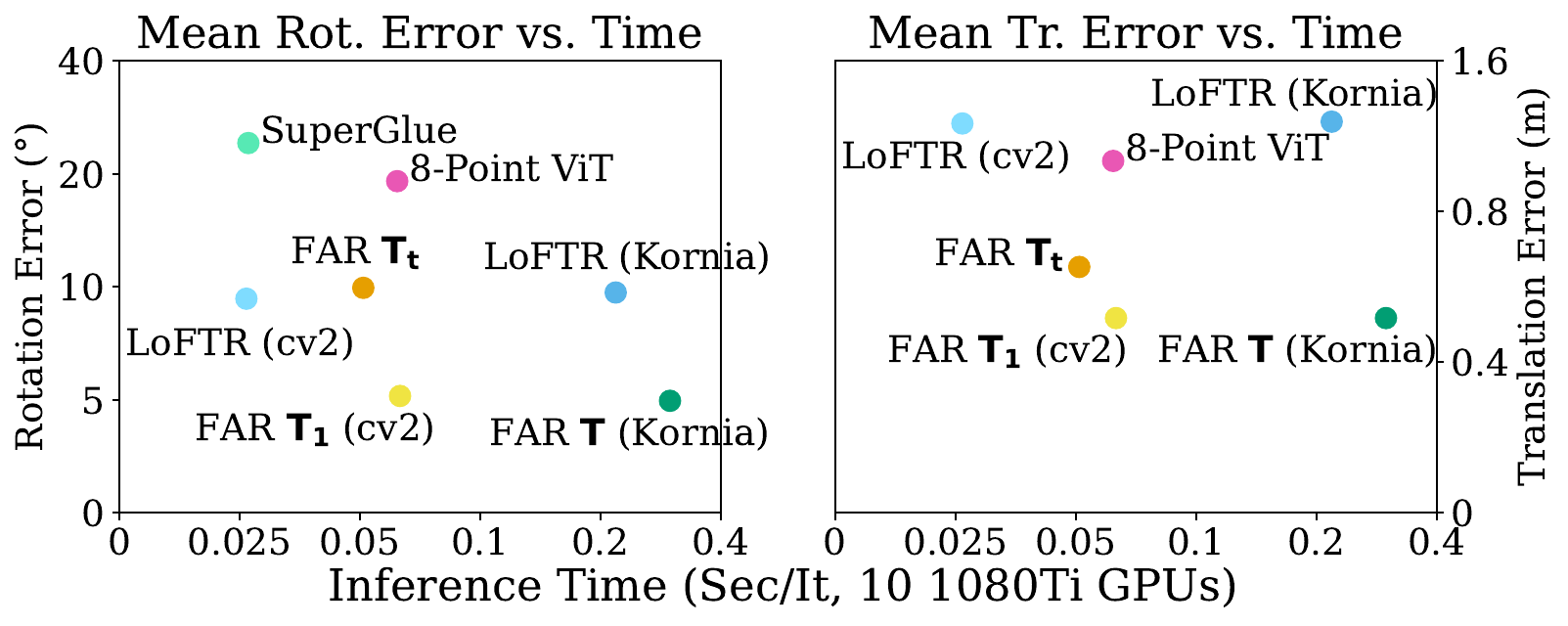}
\vspace{-1.2em}
\caption{\textbf{Efficiency on Matterport (Log Scale vs. Log Scale).} The efficient frontier includes LoFTR+Solver, $\TB_1$ and $\TB$.
FAR $\TB_t$ is strictly better than 8-Point ViT. 
FAR $\TB_1$ cuts error almost in half for little time cost. 
$\TB$ improves further, but is slower due to Prior-Guided RANSAC Kornia implementation. 
Kornia similarly slows down LoFTR+Solver. 
$\TB_1$ with Kornia (not pictured) is also worse than $\TB$, while being slower.
}
\label{fig:supp_efficiency}
\end{center}
\vspace{-1.2em}
\end{figure}

We plot error vs. time in Figure~\ref{fig:supp_efficiency}. FAR is on the efficient frontier (down and left), though it is slower than LoFTR+Solver using OpenCV (cv2). We note a Prior-Guided RANSAC implementation in cv2 could speed FAR up towards 15fps (e.g. T1).

\subsection{Random Results} 
Results on random examples are presented in Figures~\ref{fig:additional_mp3d}-~\ref{fig:additional_sl}.

The comparisons are to the same baselines as in Figure~\ref{fig:datasets}. C+S is an abbreviation for ``Correspondence Estimation + Solver", specifically LoFTR with a solver, and learned scale if necessary. 
L-B is an abbreviation for ``Learning-Based", in practice we use LoFTR with an 8-Point ViT head for Matterport3D (specifically, this is equivalent to FAR: Transformer $\TB_t$), we use 8-Point ViT for InteriorNet and StreetLearn, and use 6D Reg for Map-free Relocalization.
We choose these learning-based and correspondence-based comparisons as they are the state of the art and we build upon them for our method: on all datasets, we use LoFTR to extract correspondence; on Matterport3D, we use LoFTR for features, on InteriorNet and StreetLearn we use 8-Point ViT for features, and on Map-free Relocalization we use 6D Reg for features.

Random results give visual grounding to quantitative results from Section~\ref{sec:experiments} and are consistent with conclusions that FAR outperforms both C+S and L-B.  
Only 14 results are presented on each dataset, meaning the sample size is small, and conclusions should not be drawn from aggregate numbers. 
Rather, these examples are intended to be indicative of performance on a sample-by-sample basis.

For instance, on Matterport3D, FAR is best 10 out of 14 times in rotation and 7 out of 14 times in translation error. 
In addition, when it is not best, it typically is better than one of C+S or L-B and is typically not significantly worse than the best method.
The two qualities that it is often best and rarely worst is in line with significantly better performance than prior work averaged over a full test set.

Random Map-free Relocalization results also agree with conclusions from Section~\ref{sec:experiments} that FAR is strong.
FAR has best rotation and translation 7 out of 14 times; while rival L-B wins 2 times in rotation and 3 times in translation; C-S wins 5 times in rotation and 7 times in translation.
Despite strong performance some of the time, recall from Section~\ref{sec:experiments} C-S error is significantly higher on average than FAR.
This is showcased in the random results: when C-S fails, it does so spectacularly, for instance with rotation error of at least 120 degrees on 5 occasions, compared to 0 for FAR.
To summarize, in line with quantitative results from Section~\ref{sec:experiments}, in random samples FAR tends to be significantly more robust than C-S, while producing best results frequently. 
L-B is also more robust than C-S, but rarely produces best results.

Random results on InteriorNet also are consistent with the paper's findings.
FAR has lowest error in 7 of 14 occasions, vs. 5 for L-B and 6 for C+S.
However, the highest error for FAR is 4.2$^{\circ}$, while L-B hits 4.9$^{\circ}$ and C-S has 14.5$^{\circ}$.
On StreetLearn, FAR has a maximum error of 4.4$^{\circ}$, while L-B has errors up to 8.2$^{\circ}$ and C+S has errors of 124$^{\circ}$ and 177$^{\circ}$.
FAR has lowest error in 8 instances, vs. 3 for C+S and 5 for L-B.
When FAR beats L-B, it is often better by multiple degrees (up to 6, Figure~\ref{fig:additional_sl}, bottom left), while when L-B bests FAR, it is typically by less than three degrees.
To summarize, random results elucidate FAR is both precise and robust.

Note results on Map-free Relocalization here, as well as Figure~\ref{fig:datasets}, are on the validation set, since the test set ground truth is private -- test results are available from submitting predictions through the Map-free Relocalization website (https://research.nianticlabs.com/mapfree-reloc-benchmark/submit).
Otherwise we use test sets for random results.

\subsection{Failure Cases} 

\begin{figure*}[t]
    \centering
    \includegraphics[width=\linewidth]{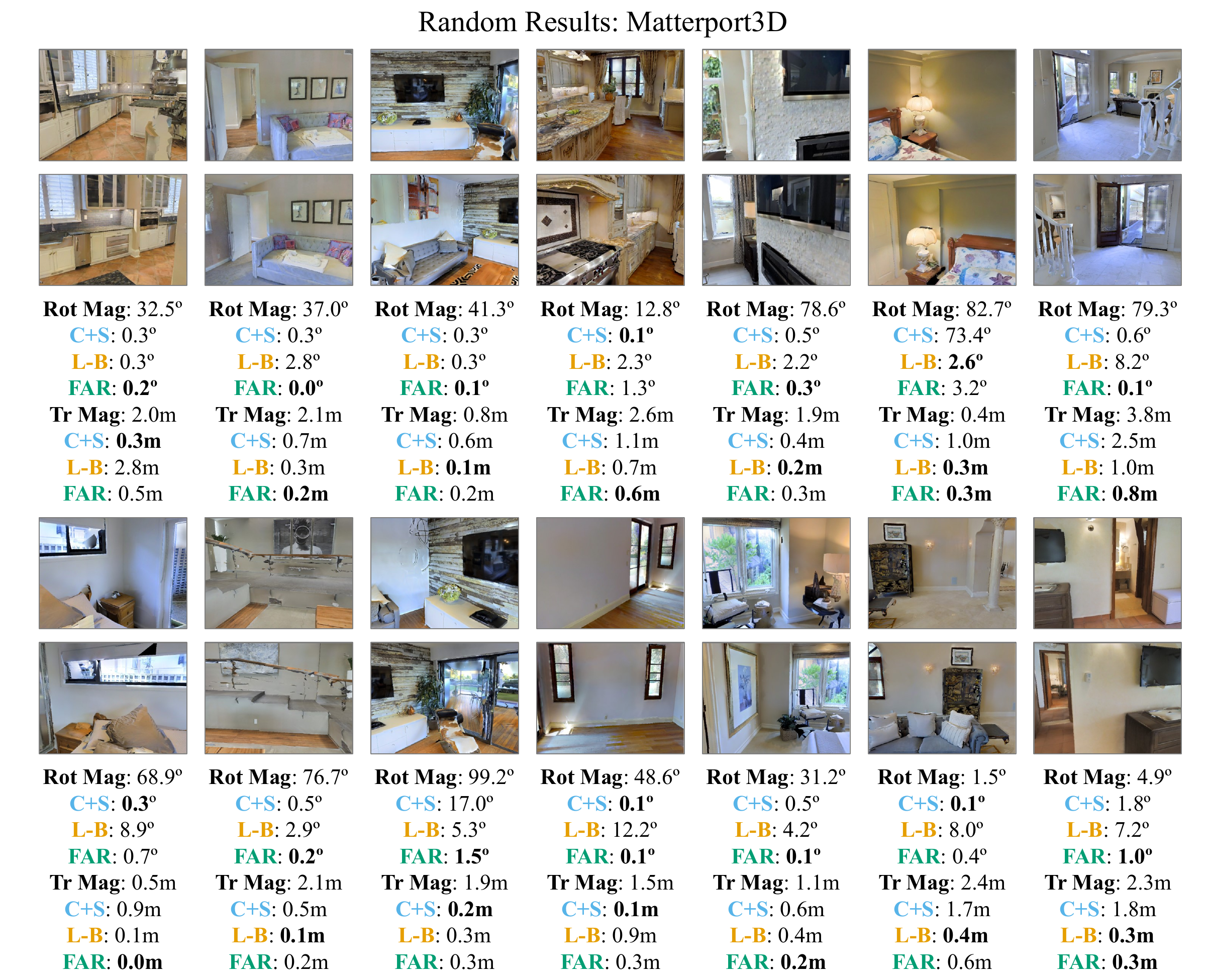}
    \vspace{-1.2em}
    \caption{\textbf{Random results on Matterport3D}. C+S: LoFTR~\cite{sun2021loftr} + Solver. L-B: LoFTR + 8-Point ViT~\cite{rockwell2022}. FAR: uses LoFTR features and correspondences. FAR is typically best. When not best, it is usually better than one of C+S or L-B.}
    \label{fig:additional_mp3d}
    \vspace{-1.2em}
\end{figure*}

\begin{figure*}[t]
    \centering
    \includegraphics[width=\linewidth]{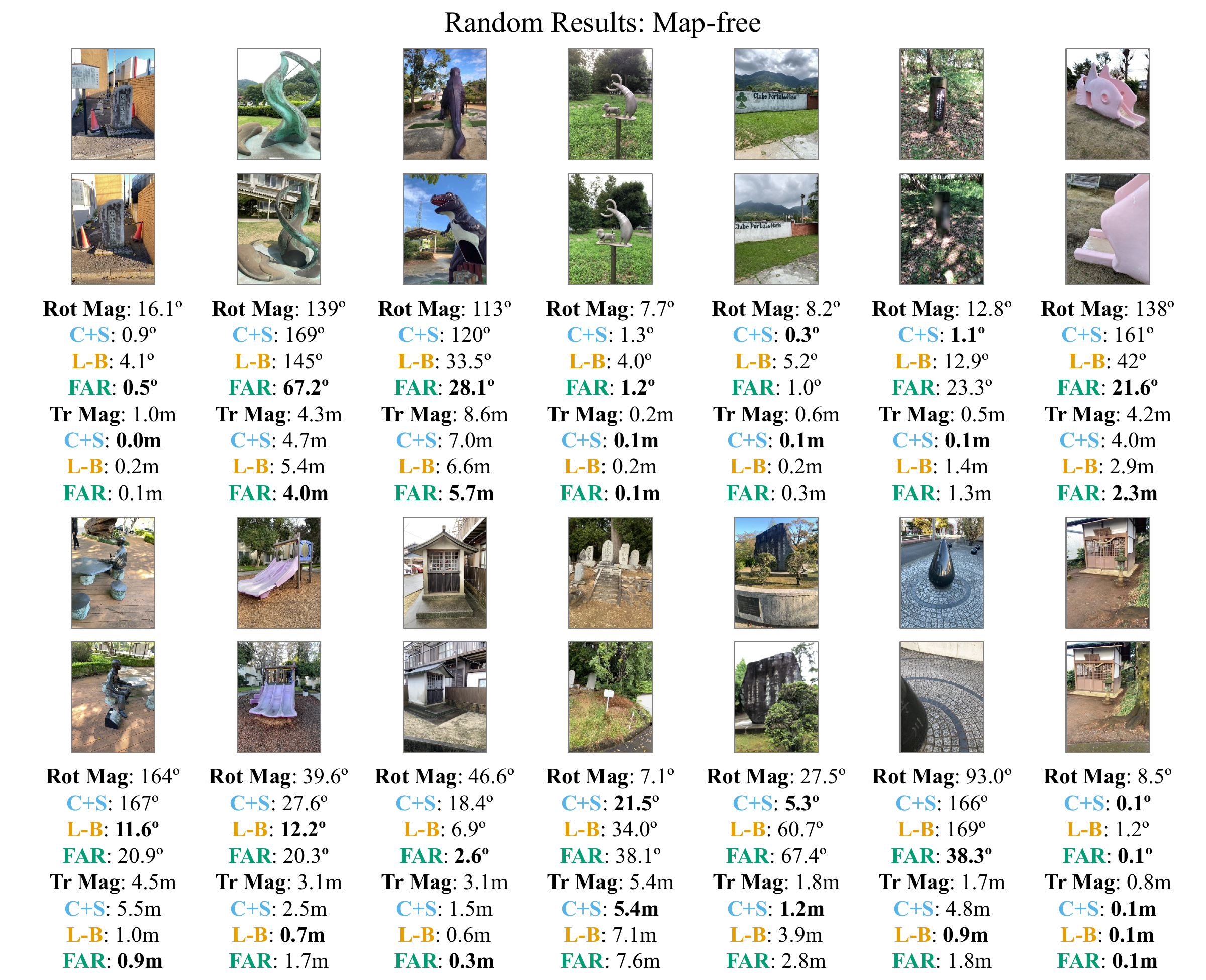}
    \vspace{-1.2em}
    \caption{\textbf{Random results on Map-free Relocalization}. C+S: LoFTR + Solver. L-B: 6D Reg~\cite{arnold2022map}. FAR: uses 6D Reg features and LoFTR correspondences.
    FAR is often best, having minimum rotation error 7 instances vs. 5 for C+S and 2 for L-B, and minimum translation error 7 times vs. 7 for C+S and 3 for L-B.
    C-S has far worse errors in failure cases than FAR (e.g. row 1 column 7).
    }
    \label{fig:additional_mapfree}
    \vspace{-1.2em}
\end{figure*}

\begin{figure*}[t]
    \centering
    \includegraphics[width=\linewidth]{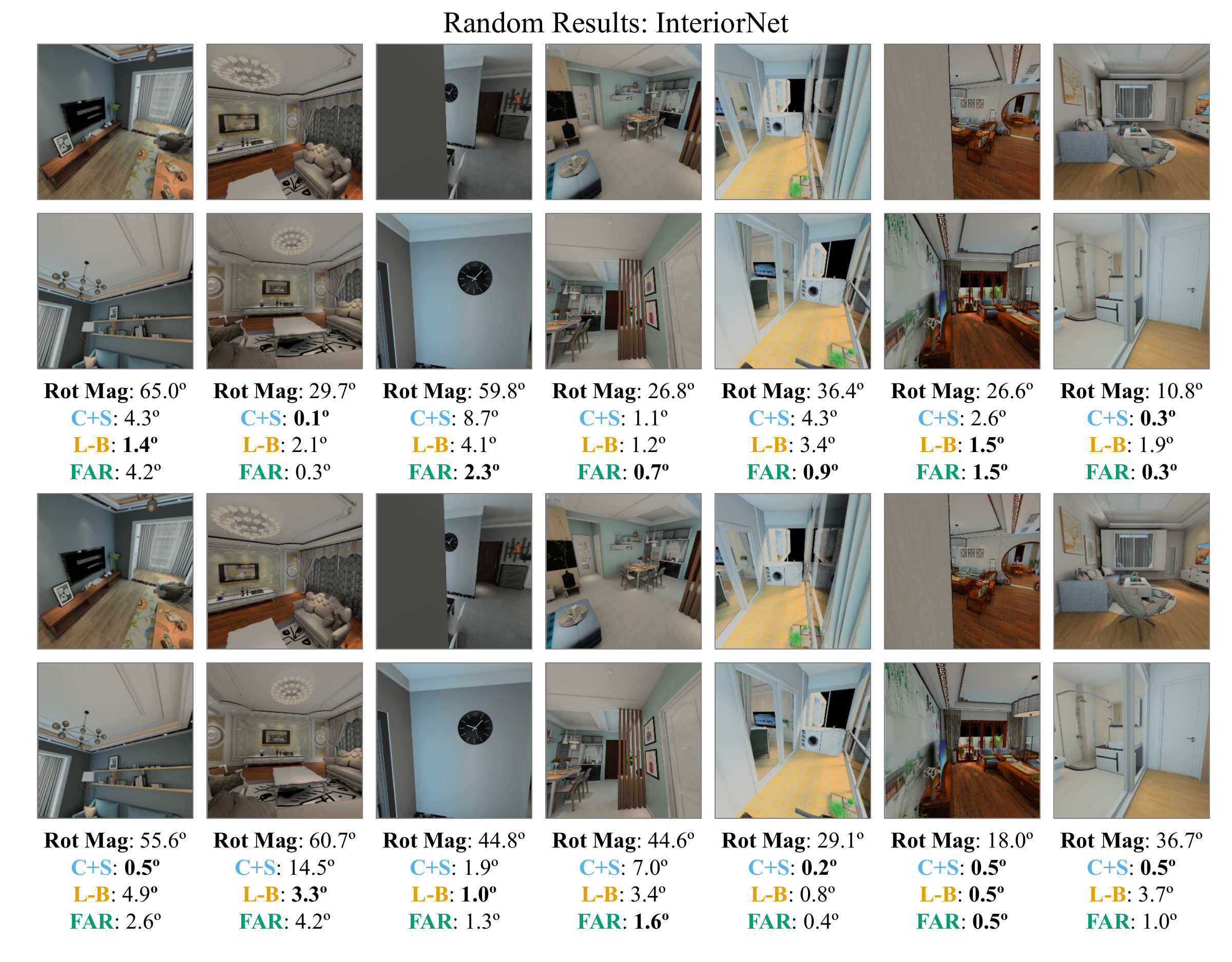}
    \vspace{-1.2em}
    \caption{\textbf{Random results on InteriorNet}. C+S: LoFTR + Solver. L-B: 8-Point ViT. FAR: uses 8-Point ViT features and LoFTR correspondences.
    FAR has the lowest error more frequently than alternatives, and has the lowest maximum error: 4.2$^{\circ}$ vs. 4.9$^{\circ}$ for L-B and 14.5$^{\circ}$ for C-S.
    }
    \label{fig:additional_in}
    \vspace{-1.2em}
\end{figure*}

\begin{figure*}[t]
    \centering
    \includegraphics[width=\linewidth]{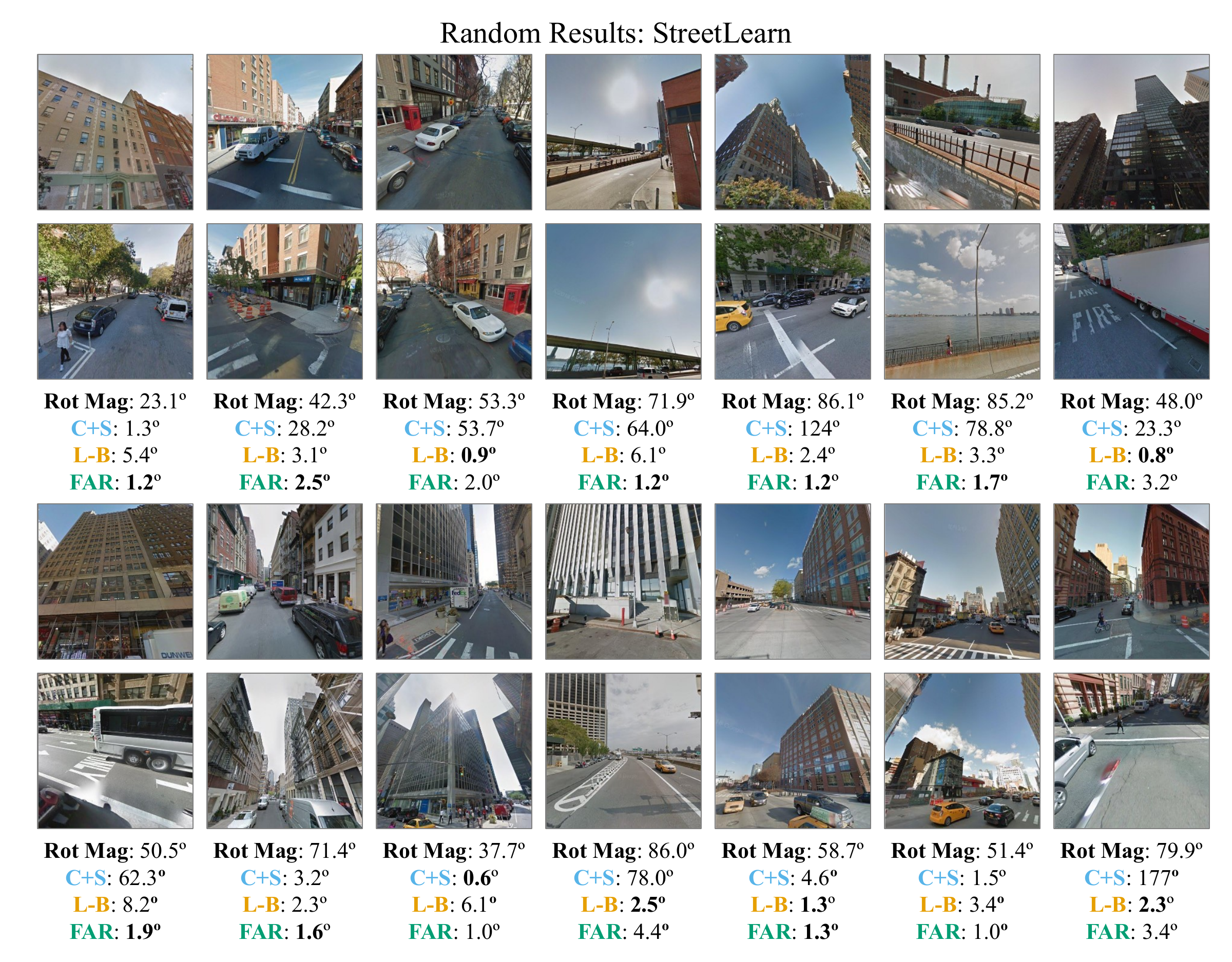}
    \vspace{-1.2em}
    \caption{\textbf{Random results on StreetLearn}. C+S: LoFTR + Solver. L-B: 8-Point ViT. FAR: uses 8-Point ViT features and LoFTR correspondences.
    FAR often has the lowest error -- here 8 times vs. 1 for C+S and 5 for L-B; and is more robust than alternatives: FAR has maximum error of 4.4$^{\circ}$, L-B has maximum error of 8.2$^{\circ}$, C+S has errors of 124$^{\circ}$ and 177$^{\circ}$.
    When FAR beats L-B, it is often better by multiple degrees (up to 6), while when L-B bests FAR, it is typically by less than three degrees.
    }
    \label{fig:additional_sl}
    \vspace{-1.2em}
\end{figure*}

We can consider some failures in the random examples from Figures~\ref{fig:additional_mp3d}-\ref{fig:additional_sl}.
For instance, some examples in Map-free are beyond the capacity of all the tested models: row 1 column 2 has all models with error above 60$^{\circ}$.
This is a case of a large rotation around a symmetric and unusually-shaped object, so much so it might be initially challenging to a human.
This is a case where recent work focused on visual disambiguation~\cite{cai2023doppelgangers} could be of assistance.

Occasionally, FAR also produces the worst results compared to C+S and L-B, for instance in Map-free results row 2 column 4.
Ideally, it would perform at least as well as the best of C+S and L-B on any instance, but this is evidence it is not always better than one alternative.

\end{appendices}

\end{document}